\documentclass[11pt,letterpaper]{article}
\usepackage{labreport}
\reportlab{}
\reporttype{Technical Report}
\reportnumber{TR-2026-01}
\reportdate{October 2026}
\reportcolors{183153}{167D8D}
\reportlinkcolor{2455A4}

\usepackage{wrapfig}
\usepackage{needspace}

\titleformat{\paragraph}[runin]
  {\normalsize\bfseries\color{black}}{}{0pt}{}

\title{LinMorph: Learning to Morph Weights for Efficient Video DiT Linearization}

\author{%
  Lingyi Long\textsuperscript{1},
  Congyue Zhao\textsuperscript{1},
  Xiaoou Li\textsuperscript{1},
  Yanmin Qian\textsuperscript{1,2},
  Tao Huang\textsuperscript{1,2$\dagger$}%
}

\reportaffiliations{%
  \textsuperscript{1}Shanghai Jiao Tong University\qquad \textsuperscript{2}VUI Labs Research
}

\reportauthornote{$\dagger$Correspondence: \href{mailto:t.huang@sjtu.edu.cn}{t.huang@sjtu.edu.cn}.}

\reportresourcegroup{Project and Code}{%
  \reportresourceitem
    {Project page}
    {https://lonelyling.github.io/LinMorph/}
    {\nolinkurl{https://lonelyling.github.io/LinMorph/}}
  \reportresourceitem
    {Code repository}
    {https://github.com/lonelyling/LinMorph}
    {\nolinkurl{https://github.com/lonelyling/LinMorph}}
  \reportresourceitem
    {Model checkpoints}
    {https://huggingface.co/Lonelly/LinMorph}
    {\nolinkurl{https://huggingface.co/Lonelly/LinMorph}}
}

\begin{document}

\maketitle
\raggedbottom

\begin{abstract}
Post-hoc linearization is an attractive way to accelerate pretrained Video
Diffusion Transformers (Video DiTs), yet existing methods largely treat the
quality loss induced by attention replacement as a recovery problem. We show
that a more fundamental bottleneck lies upstream: \emph{initialization}.
Random initialization discards pretrained knowledge, whereas manual weight
reuse transfers it through rigid and incomplete source--target correspondences.
We introduce \textbf{LinMorph}, which recasts Video DiT linearization as a
learnable softmax-to-linear weight transformation problem. Instead of copying
predefined counterparts, LinMorph learns a cross-parameter morphism that lets
each target Gated DeltaNet (GDN) parameter draw jointly from pretrained query,
key, value, and output projections, including for parameters with no direct
softmax counterpart. A single morphism is shared across layers and learned
through local teacher alignment, while producing layer-specific
initializations from their respective pretrained weights. The resulting
weights are then materialized as independent GDN modules for progressive
distillation and joint recovery.
Since the morphism operates in weight space rather than token space, it can be
learned on short, low-resolution clips and directly reused at larger video
scales.
Across pretrained Video DiTs, LinMorph enables markedly faster recovery and
better final generation quality than random initialization and heuristic
weight reuse. Results show that efficient linearization depends not only
on how a converted model is recovered, but critically on how pretrained
knowledge is transferred before recovery.
\end{abstract}

\setcounter{tocdepth}{2}
\reportcontents

\section{Introduction}
Video diffusion transformers (Video DiTs)~\citep{ma2024latte,yang2024cogvideox,kong2024hunyuanvideo,wan2025} have recently emerged as a dominant paradigm for high-quality video generation, benefiting from the scalability and expressive capacity of Transformer architectures~\citep{peebles2023scalable}. However, their computational cost grows rapidly with the number of spatio-temporal tokens, making full softmax attention increasingly expensive for high-resolution and long-duration video generation. Linear attention provides a promising alternative by reducing the quadratic complexity of token interaction to linear or near-linear complexity~\citep{katharopoulos2020transformers,yang2024gla,yang2024deltanet}. Yet, training a linear-attention Video DiT from scratch is itself prohibitively expensive. A more practical route is therefore to \emph{linearize} an existing pretrained Video DiT by replacing its softmax attention modules with efficient linear counterparts, followed by a short recovery stage to restore generation quality~\citep{ghafoorian2025attention,huang2025linvideo,ghafoorian2026rehyat}. This post-hoc linearization paradigm enables efficient video generation while preserving the value of expensive pretrained models.

Despite its appeal, post-hoc linearization faces an overlooked initialization
problem. Existing approaches typically either initialize the new linear
attention from scratch and rely on recovery training to reconstruct the lost
functionality~\citep{wang2025lit}, or manually copy seemingly corresponding
softmax projections into the target operator~\citep{
kasai2021finetuning,wang2025lit}. The former discards knowledge already encoded
in the pretrained attention weights, while the latter relies on rigid and
often incomplete source--target correspondences. This limitation becomes
particularly pronounced for modern recurrent linear attention operators, whose
parameterizations contain components with no direct softmax counterpart.
Consequently, a substantial difficulty of Video DiT linearization
may arise before recovery begins: from how the replacement module is
initialized.

This observation suggests a different formulation. Rather than asking
\emph{which pretrained softmax weights should be copied to which
linear-attention parameters}, we ask \emph{how pretrained softmax weights
should be transformed into an effective linear-attention initialization}.
From this perspective, manual copying is only a highly constrained
transformation with a predefined one-to-one correspondence. A learned
transformation can instead exploit information jointly distributed across
multiple pretrained projections and construct parameters even when no explicit
counterpart exists in the source attention.

\begin{figure}[t]
    \centering
    \includegraphics[width=0.98\linewidth]{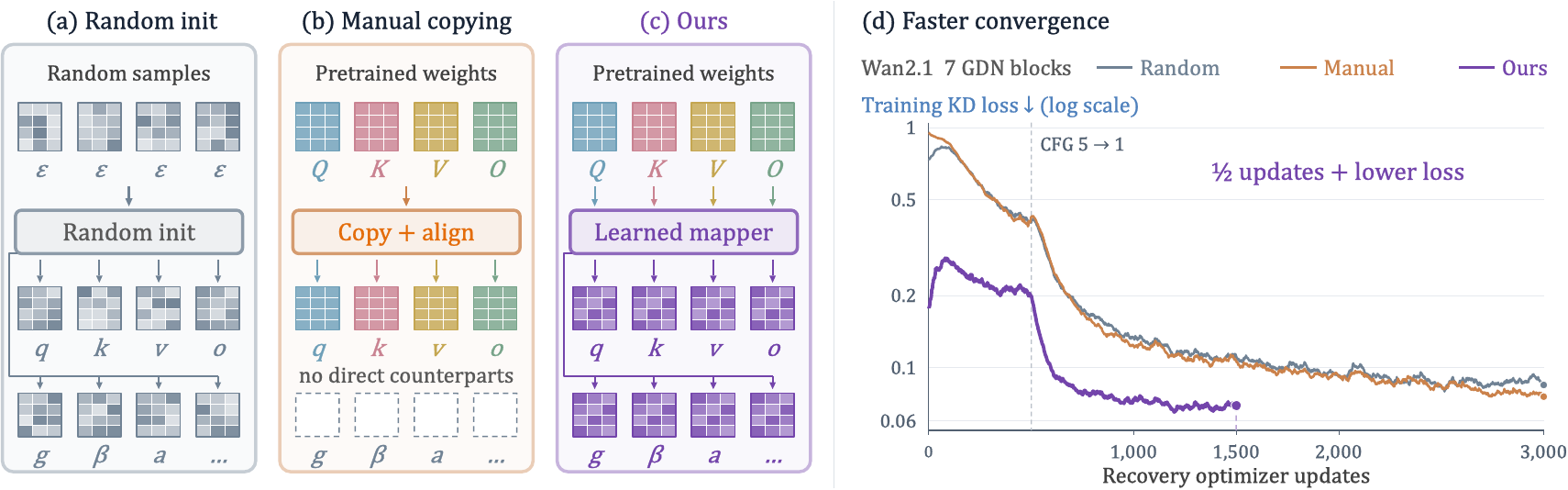}
    \caption{
    \textbf{Comparisons of different weight initialization methods.} (a) Random initialization discards pretrained knowledge, while (b) manual copying relies on partial one-to-one correspondences. (c) Our LinMorph learns a cross-parameter transformation that initializes both matched and target-specific linear-attention parameters. (d) The resulting initialization yields substantially faster recovery and lower KD loss.
    }
    \label{fig:linmorph_motivation}
\end{figure}

Building on this idea, we propose \textbf{LinMorph}, a parameter-efficient
weight-morphing framework for post-hoc Video DiT linearization.
LinMorph learns a source-conditioned transformation shared across layers:
each layer provides its own pretrained softmax weights, while the same learned
rule constructs its linear-attention initialization. The transformation is
fitted through local teacher alignment in a compact shared parameter space,
after which the generated weights are materialized as independent replacement
modules and optimized by progressive distillation and joint recovery.
Because the morphism acts on model weights rather than video tokens, it can be
learned from short, low-resolution clips and reused to initialize recovery at
larger video scales. Figure~\ref{fig:linmorph_motivation} contrasts LinMorph
with random initialization and manual weight reuse.

Experiments on pretrained Video DiTs show that LinMorph provides a
substantially stronger starting point than both random initialization and
heuristic weight reuse, leading to faster recovery and better final generation
quality. The learned morphism also transfers across video resolutions and
durations without mapper refitting, allowing the initialization to be learned
at a cheaper video scale and reused for more expensive recovery. These results
highlight a complementary axis for efficient Video DiT linearization:
beyond designing a better attention operator or recovery objective, it is
equally important to determine how pretrained softmax knowledge is transferred
into the new parameterization.

\section{LinMorph: Softmax-to-Linear Weight Morphism}
\label{sec:linmorph}
\begin{figure}[t]
    \centering
    \includegraphics[width=0.95\linewidth]{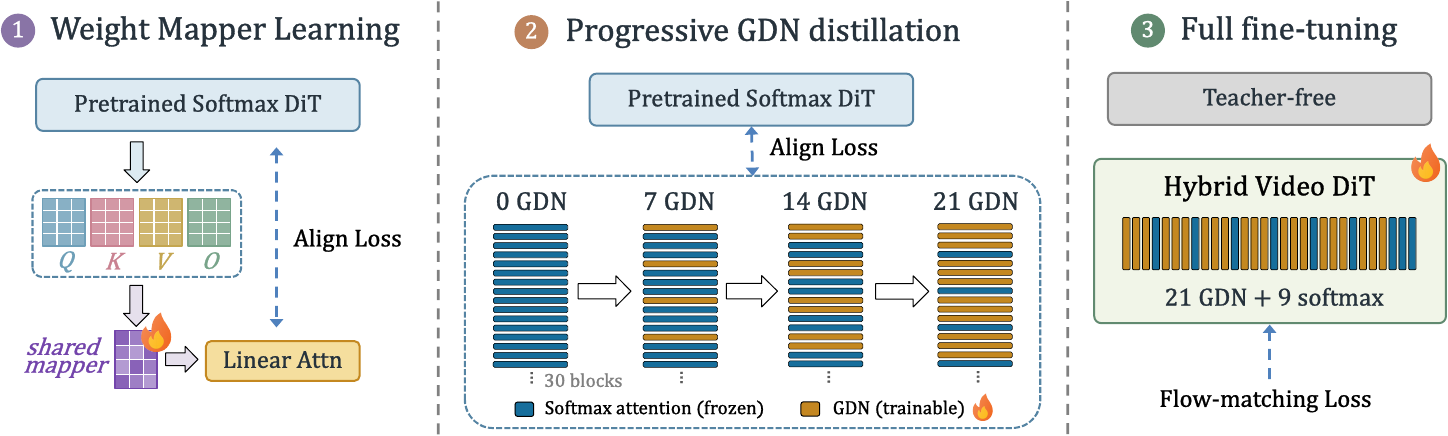}
    \caption{\textbf{Overview of LinMorph.}
A shared weight morphism is first learned from the frozen softmax teacher and
used to generate GDN initialization weights. After materialization, the mapper
is discarded and the GDN modules are progressively introduced and recovered
by prediction distillation. The resulting hybrid Video DiT is finally
fine-tuned end-to-end without the teacher.}
    \label{fig:linmorph_pipeline}
\end{figure}

LinMorph learns a source-conditioned transformation shared across
layers to initialize Gated DeltaNet (GDN) replacements
in pretrained Video DiTs~\citep{yang2024gateddeltanet}.
Each layer supplies its own softmax weights, while the shared rule
constructs its GDN initialization.
Figure~\ref{fig:linmorph_pipeline} summarizes three stages:
local mapper fitting, progressive distillation of the materialized
replacements, and teacher-free joint fine-tuning.

\subsection{Cross-Parameter Weight Transformation}
\label{sec:cross_parameter}

\begin{wrapfigure}{r}{0.35\textwidth}
    \centering
    \includegraphics[width=\linewidth]{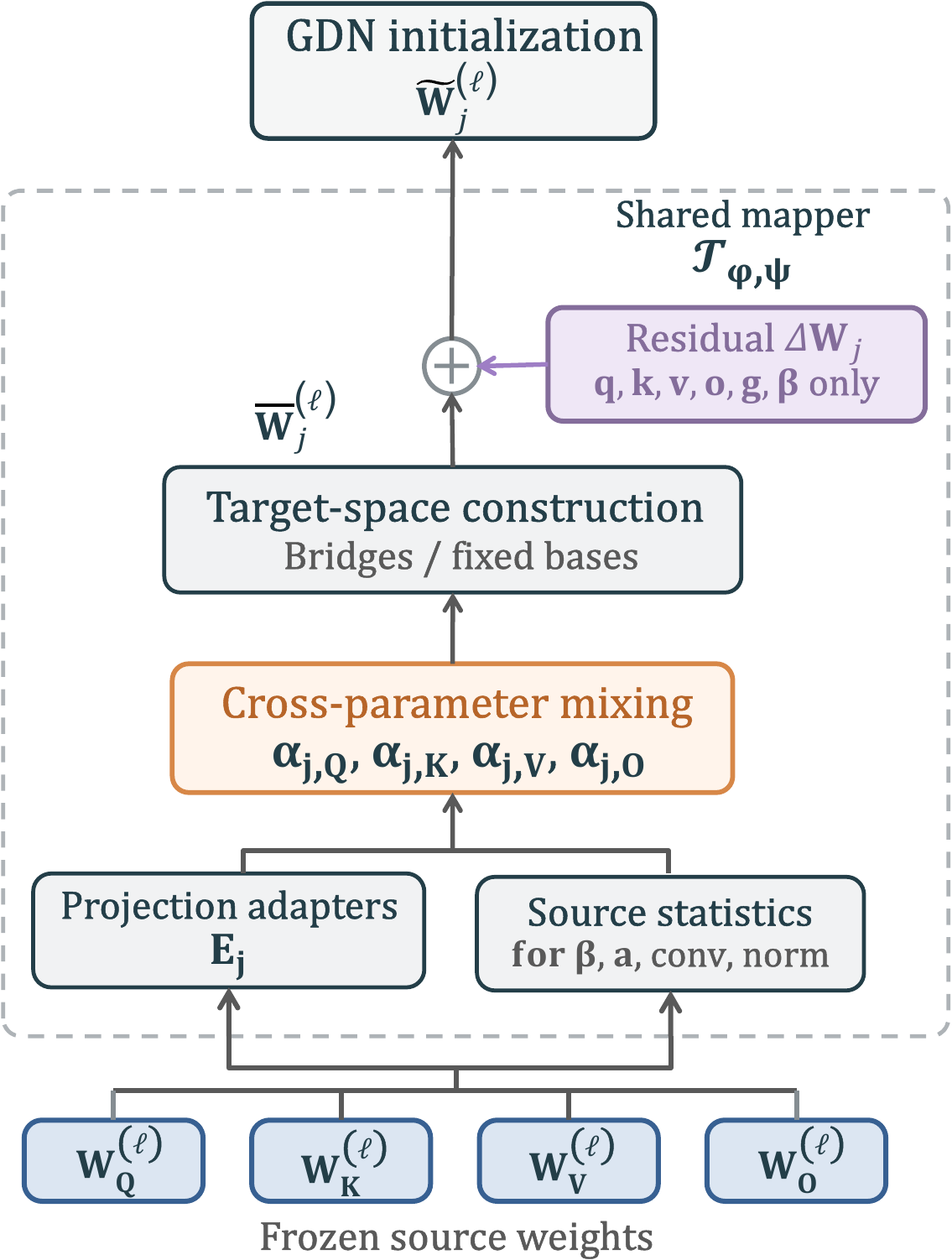}
    \caption{\textbf{Shared weight mapper.}
    Shown for a target parameter $j$.
    Alignment and construction are target-dependent;
    residual correction applies only to $q,k,v,o,g,\beta$.}
    \label{fig:model_structure}
\end{wrapfigure}

For layer $\ell$, let
$\Theta_{\mathrm{soft}}^{(\ell)}$
denote the pretrained softmax-attention parameters and
$\widetilde{\Theta}_{\mathrm{lin}}^{(\ell)}$
the initialization of its GDN replacement. LinMorph applies a weight morphism
shared across participating layers,
\begin{equation}
\widetilde{\Theta}_{\mathrm{lin}}^{(\ell)}
=
\mathcal{T}_{\phi,\psi}
\left(
\Theta_{\mathrm{soft}}^{(\ell)}
\right),
\end{equation}
where $\phi$ parameterizes a source-conditioned base transformation and
$\psi$ a residual correction. The transformation parameters are shared, but
the resulting initializations are layer-specific because each layer provides
different pretrained source weights. LinMorph therefore shares the
\emph{transformation rule}, rather than the weights of the resulting GDN
modules. Here, ``morphism'' refers to a learned cross-architecture weight
transformation, not an exact function-preserving mapping between softmax
attention and GDN. Figure~\ref{fig:model_structure} illustrates the shared mapper for an individual target parameter.

\paragraph{Cross-parameter mixing.}
Rather than prescribing a one-to-one source--target correspondence, LinMorph
allows each target parameter to draw information jointly from all pretrained
softmax projections. Let
$\mathbf{W}_{s}^{(\ell)}\in\mathbb{R}^{D\times D}$,
$s\in\mathcal{S}=\{Q,K,V,O\}$,
denote the four source projections. Given shape-aligned source matrices
$\{\mathbf{Z}_{s}\}_{s\in\mathcal{S}}$, we define
\begin{equation}
\mathcal{M}_{j}
\left(
\{\mathbf{Z}_{s}\}_{s\in\mathcal{S}}
\right)
=
\sum_{s\in\mathcal{S}}
\alpha_{j,s}\mathbf{Z}_{s},
\end{equation}
where $\alpha_{j,s}$ are learned unconstrained coefficients shared across
layers. Each target parameter $j$ thus learns not only \emph{which} pretrained
projections are useful, but also \emph{how} their information should be
combined.

\paragraph{Dimension and parameter alignment.}
Softmax attention and GDN need not have matching internal dimensions. We
introduce target-specific adapters $\mathbf{E}_{j}$ for the attention-channel
dimension, together with hidden-width bridges
$\mathbf{B}_{\mathrm{up}}$ and $\mathbf{B}_{\mathrm{down}}$
between the pretrained DiT width $D$ and the GDN width $D_g$.
For the main GDN projections, the source-conditioned mixtures are
\begin{align}
\mathbf{W}_{j,\mathrm{mix}}^{(\ell)}
&=
\mathcal{M}_{j}
\left(
\{\mathbf{E}_{j}\mathbf{W}_{s}^{(\ell)}\}_{s\in\mathcal{S}}
\right),
&& j\in\{q,k,v,g\}, \\
\mathbf{W}_{o,\mathrm{mix}}^{(\ell)}
&=
\mathcal{M}_{o}
\left(
\{\mathbf{W}_{s}^{(\ell)}\mathbf{E}_{o}\}_{s\in\mathcal{S}}
\right),
\end{align}
from which the base target weights are constructed as
\begin{equation}
\overline{\mathbf{W}}_{j}^{(\ell)}
=
\mathbf{W}_{j,\mathrm{mix}}^{(\ell)}
\mathbf{B}_{\mathrm{down}},
\quad j\in\{q,k,v,g\},
\qquad
\overline{\mathbf{W}}_{o}^{(\ell)}
=
\mathbf{B}_{\mathrm{up}}
\mathbf{W}_{o,\mathrm{mix}}^{(\ell)}.
\end{equation}
The bridges connect the GDN replacement to the surrounding DiT while
preserving the external hidden width. GDN additionally contains parameters without direct softmax counterparts,
including update and decay projections, short convolutions, and normalization
parameters. We construct these parameters from compact source-conditioned
statistics of the pretrained projections and apply the same learned
cross-parameter mixing principle. This allows pretrained information to
contribute even when an explicit source--target correspondence does not exist.
Detailed constructions are provided in
Appendix~\ref{app:linmorph_auxiliary}.

\paragraph{Residual correction.}
The source-conditioned transformation provides a structured base
initialization, but need not fully account for the parameterization gap between
softmax attention and GDN. We therefore augment each generated weight with a
shared residual in the native target space,
\begin{equation}
\label{eq:linmorph_residual}
\widetilde{\mathbf{W}}_{j}^{(\ell)}
=
\overline{\mathbf{W}}_{j}^{(\ell)}
+
\Delta\mathbf{W}_{j},
\qquad
j\in\{q,k,v,o,g,\beta\},
\end{equation}
where $\widetilde{\mathbf{W}}_{j}^{(\ell)}$ denotes the final weight used to
initialize the replacement before recovery. The mixing coefficients, adapters,
bridges, and residual corrections are shared across participating layers,
while the source-conditioned term preserves layer-specific information.

With matching dimensions, identity adapters and bridges, zero residuals, and a
one-hot source mixture, LinMorph reduces to direct one-to-one weight copying.
Manual reuse can therefore be viewed as a restricted special case of the
proposed transformation, whereas LinMorph learns how pretrained information
should be reassigned to the target parameterization.

\subsection{Shared Morphism Optimization}
\label{sec:learning_morphism}

\paragraph{Local teacher alignment.}
We optimize the shared morphism by locally matching each mapped GDN
replacement to its corresponding pretrained softmax block. Given a noisy video
latent $\mathbf{z}_{t}$ and text condition $c$, a frozen teacher forward pass
provides, for each participating layer $\ell$, the block input
$\mathbf{X}_{T}^{(\ell)}$, the attention output
$\mathbf{A}_{T}^{(\ell)}$, and the complete block output
$\mathbf{Y}_{T}^{(\ell)}$.

For each layer, we then replay the block using the mapped GDN initialization
$\widetilde{\Theta}_{\mathrm{lin}}^{(\ell)}$, while fixing its input to the
teacher activation $\mathbf{X}_{T}^{(\ell)}$. This produces
$\widehat{\mathbf{A}}^{(\ell)}$ and
$\widehat{\mathbf{Y}}^{(\ell)}$ for the mapped attention module and the full
block, respectively. All non-attention components remain frozen, while
gradients are propagated through them to the shared morphism. Using the
teacher activation independently at every layer isolates the local conversion
error and prevents mismatches from preceding replacements from accumulating
during mapper optimization.

We match both the attention output and its effect on the surrounding block:
\begin{equation}
\label{eq:linmorph_map_loss}
\mathcal{L}_{\mathrm{map}}
=
\frac{1}{|\mathcal{B}|}
\sum_{\ell\in\mathcal{B}}
\left[
\mathcal{D}
\left(
\widehat{\mathbf{A}}^{(\ell)},
\mathbf{A}_{T}^{(\ell)}
\right)
+
\mathcal{D}
\left(
\widehat{\mathbf{Y}}^{(\ell)},
\mathbf{Y}_{T}^{(\ell)}
\right)
\right],
\end{equation}
where $\mathcal{D}(\cdot,\cdot)$ denotes element-wise mean squared error.
The first term directly aligns the replacement operator, while the second
measures whether the mapped attention remains compatible with the surrounding
pretrained block.

\paragraph{Two-stage optimization.}
We first optimize the source-conditioned base transformation $\phi$ with the
residual correction fixed to zero. We then freeze $\phi$ and optimize only
$\psi$ under the same alignment objective. This decomposition first captures
the reusable transformation structure shared across layers, and then corrects
the remaining mismatch directly in the native GDN parameter space.

\paragraph{Materialization.}
After optimization, we evaluate
$\mathcal{T}_{\phi^{\star},\psi^{\star}}$
once on the pretrained parameters of each participating layer and materialize
the resulting
$\widetilde{\Theta}_{\mathrm{lin}}^{(\ell)}$
as independent GDN parameters. The shared mapper is then discarded; the
materialized modules are no longer tied across layers and can be optimized
independently during subsequent recovery.

Because the morphism is shared across layers and operates in weight space,
mapper fitting optimizes a much smaller parameter set than directly recovering
all converted GDN modules. Moreover, its parameterization depends on model
dimensions rather than video token count at a fixed architecture. We therefore
fit the morphism using short, low-resolution clips and reuse the resulting
initialization for recovery at larger video scales; this cross-scale behavior is examined in Appendix~\ref{app:cross_scale}.

\subsection{Progressive Conversion and Recovery}
\label{sec:linmorph_recovery}

\paragraph{Progressive distillation.}
The materialized LinMorph weights provide the starting point for converting
the pretrained Video DiT. Rather than replacing many softmax blocks at once,
we progressively increase the number of GDN replacements, e.g.,
$0\!\rightarrow\!7\!\rightarrow\!14\!\rightarrow\!21$ for Wan2.1, as
illustrated in Figure~\ref{fig:linmorph_pipeline}. At each stage, newly
introduced GDN modules are initialized by LinMorph, while previously converted
modules inherit their recovered weights. The converted modules and their
bridges are then adapted by matching the velocity prediction of the frozen
softmax teacher,
\begin{equation}
\mathcal{L}_{\mathrm{KD}}
=
\mathrm{MAE}
\left(
\mathbf{v}_{\theta}(\mathbf{z}_t,t,c),
\mathbf{v}_{T}(\mathbf{z}_t,t,c)
\right).
\end{equation}
Progressive conversion limits the architectural change introduced at each
stage, allowing the mapped initialization to be adapted before further
softmax blocks are replaced.

\paragraph{Teacher-free joint fine-tuning.}
Once the target replacement ratio is reached, we remove teacher supervision
and jointly fine-tune the entire hybrid Video DiT with
flow-matching objective,
\begin{equation}
\mathcal{L}_{\mathrm{FM}}
=
\mathbb{E}
\left[
w(t)\,
\mathrm{MSE}
\left(
\mathbf{v}_{\theta}(\mathbf{z}_t,t,c),
\boldsymbol{\epsilon}-\mathbf{z}_0
\right)
\right].
\end{equation}
This final stage allows the softmax and GDN blocks to co-adapt under the
original generative objective, producing the recovered hybrid model used for
inference.

\section{Understanding the Learned Weight Morphism}
\label{sec:understanding}

We isolate the learned morphism before downstream recovery to understand
why it provides an effective initialization. We first examine the
cross-parameter structure discovered by LinMorph, then evaluate the
functional fidelity of the resulting initialization.

\subsection{Learned Cross-Parameter Structure}
\label{sec:learned_structure}

\begin{wrapfigure}{r}{0.5\linewidth}
    \centering
    \includegraphics[width=\linewidth, height=0.6\linewidth]{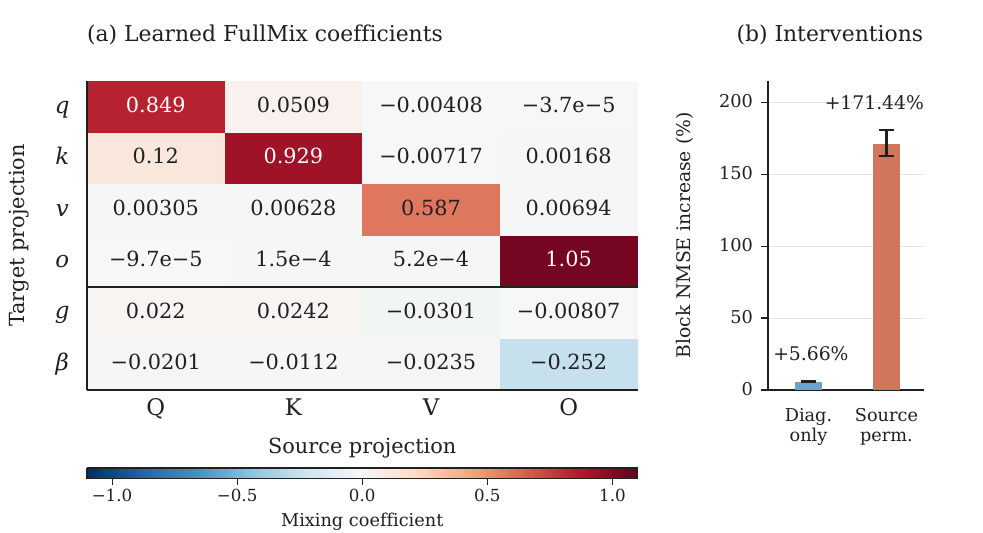}
    \caption{\textbf{Learned cross-parameter structure.} Left: source-mixing coefficients. Right: block-NMSE degradation after restricting or permuting the learned source assignments.}
    \label{fig:learned_morphism}
\end{wrapfigure}

LinMorph does not simply recover the namesake correspondences used by manual
weight copying. Figure~\ref{fig:learned_morphism} visualizes the learned
source-mixing coefficients $\alpha_{j,s}$. While the query and key targets
retain strong contributions from their corresponding source projections, they
also draw non-negligible information from other pretrained weights:
cross-projection terms account for $6.08\%$ and $12.20\%$ of their absolute
coefficient mass, respectively. For target parameters without a direct softmax
counterpart, the learned assignments are even more distinctive. For example,
$\mathbf{W}_{O}$ contributes $82.16\%$ of the absolute coefficient mass for
the update projection $\beta$. These patterns suggest that the useful
softmax-to-linear correspondence is neither strictly one-to-one nor determined
by parameter names.

We further test whether these learned correspondences are functionally
meaningful rather than incidental. Without any additional optimization, we
intervene on the fitted mapper while keeping all other parameters fixed.
Restricting $q,k,v,o$ to their namesake source projections increases block
NMSE by $5.66\%$. More strikingly, cyclically permuting the learned source
assignments for GDN-specific parameters increases block NMSE by $171.44\%$,
despite preserving the norms of their coefficient vectors. The paired
$95\%$ bootstrap intervals are $[5.34,5.99]\%$ and
$[162.73,180.64]\%$.

Together, these results show that LinMorph learns structured
cross-parameter relationships that are important for functional alignment.
The benefit therefore comes not merely from reusing pretrained weights, but
from learning how information distributed across different softmax
projections should be reassigned to the target linear-attention
parameterization.

\par
\Needspace{18\baselineskip}
\subsection{Functional Fidelity of the Initialization}
\label{sec:init_fidelity}

\begin{wraptable}{r}{0.48\textwidth}
    \centering
    \vspace{-1em}
    \caption{\textbf{Pre-recovery initialization fidelity.}
    Evaluated on 64 HQ videos, 30 blocks, and five diffusion timesteps.}
    \label{tab:init_fidelity}

    \footnotesize
    \setlength{\tabcolsep}{2pt}
    \begin{tabular}{lcc}
        \toprule
        Initialization
        & Block NMSE $\downarrow$
        & Block Cos. $\uparrow$ \\
        \midrule
        Random & 0.09483 & 0.96211 \\
        Manual copying & 0.09446 & 0.96209 \\
        \midrule
        Base & 0.01360 & 0.99321 \\
        Base + residual
        & \textbf{0.01239}
        & \textbf{0.99381} \\
        \bottomrule
    \end{tabular}
\end{wraptable}

We next evaluate the mapped initialization before any downstream recovery.
As shown in Table~\ref{tab:init_fidelity}, the base morphism already achieves
high teacher-forced block fidelity. Residual refinement further improves both
attention- and block-level alignment, reducing block NMSE from $0.01360$ to
$0.01239$ and increasing block cosine similarity from $0.99321$ to $0.99381$.

These results are measured before progressive distillation or joint
fine-tuning, showing that the learned weight transformation itself places the
GDN replacement close to the pretrained softmax block in function space.
The residual correction then captures the remaining mismatch in the target
parameterization.
Additional analyses of cross-scale reuse and cross-layer sharing are
provided in Appendices~\ref{app:cross_scale}
and~\ref{app:cross_layer}, respectively.

\section{Experiments}

\subsection{Experimental Setup}
\label{sec:experimental_setup}

\paragraph{Backbones and data.}
Our primary backbone is the 30-block Wan2.1-T2V-1.3B~\citep{wan2025},
for which we replace 14 or 21 softmax-attention blocks with GDN.
To test cross-backbone applicability, we additionally replace six blocks of
Wan2.2-TI2V-5B~\citep{wan2025wan22} in text-to-video mode.
All initialization ablations within each backbone use the same target
architecture and recovery recipe.
Wan2.1 uses our curated high-quality internet-video set
(\textit{HQ videos}) for mapper fitting and recovery,
while Wan2.2 recovery uses OpenVid~\citep{nan2025openvid}.
The Wan2.2 experiment uses a separately fitted shared-mapper variant;
its details are provided in Appendix~\ref{app:experimental_protocol}.

\paragraph{Evaluation.}
We evaluate functional fidelity with normalized mean squared error (NMSE) and
cosine similarity at both block and final-output levels. Teacher-forced metrics
isolate individual replacements on teacher hidden states, whereas depth-rollout
metrics propagate student representations through the DiT and measure the
composed model. All initialization and recovery-schedule ablations use the same
fixed HQ evaluation subset. Generation quality is evaluated with
VBench~\citep{huang2024vbench} and VBench-2.0~\citep{zheng2025vbench2}.
Detailed training settings, metric definitions, evaluation subsets, and
checkpoint accounting are provided in
Appendix~\ref{app:experimental_protocol}.

\subsection{Initialization Quality and Recovery Efficiency}
\label{sec:init_ablation}

We isolate the effect of initialization by holding the target architecture
and recovery recipe fixed, and compare LinMorph with random initialization
and dimension-aligned manual copying. Table~\ref{tab:initialization} reports functional fidelity, while
Figures~\ref{fig:linmorph_motivation}(d) and~\ref{fig:recovery_curves}
show the recovery dynamics on Wan2.1 and Wan2.2, respectively.

On Wan2.1 with seven GDN blocks, LinMorph reduces depth-rollout block NMSE by
$61.4\%$ over manual copying and $65.5\%$ over random initialization after
1,500 updates, while lowering final-output NMSE to $0.01013$.
It also outperforms random initialization trained for twice as many updates
($0.01682$ at 3,000 updates), showing that a stronger initialization improves
both intermediate alignment and end-to-end prediction while substantially
reducing recovery effort. Consistent with these fidelity gains,
Figure~\ref{fig:linmorph_motivation}(d) shows that LinMorph reaches
a lower training KD loss in 1,500 updates than either baseline
in 3,000 updates.



\begin{table}[t]
\centering
\footnotesize
\setlength{\tabcolsep}{2.2pt}
\caption{\textbf{Controlled initialization ablation on two Video DiT backbones.}
Within each backbone, all methods use the same target architecture, recovery
recipe, and evaluation protocol.}
\label{tab:initialization}

\begin{minipage}[t]{0.49\linewidth}
\centering
\textit{Wan2.1: 7 GDN blocks}
\vspace{2pt}

\begin{tabular}{@{}lrrrrr@{}}
\toprule
Init. & Upd. &
\shortstack{Block\\NMSE $\downarrow$} &
\shortstack{Block\\Cos. $\uparrow$} &
\shortstack{Output\\NMSE $\downarrow$} &
\shortstack{Output\\Cos. $\uparrow$} \\
\midrule
Random & 1,500 & 0.1522 & 0.9315 & 0.0250 & 0.9873 \\
Manual & 1,500 & 0.1359 & 0.9379 & 0.0230 & 0.9884 \\
Random & 3,000 & 0.1168 & 0.9449 & 0.0168 & 0.9915 \\
Manual & 3,000 & 0.0905 & 0.9565 & 0.0137 & 0.9931 \\
\midrule
\textbf{LinMorph} & 1,500 & \textbf{0.0525} & \textbf{0.9738} & \textbf{0.0101} & \textbf{0.9949} \\
\bottomrule
\end{tabular}
\end{minipage}
\hfill
\begin{minipage}[t]{0.49\linewidth}
\centering
\textit{Wan2.2: 6 GDN blocks}
\vspace{2pt}

\begin{tabular}{@{}lrrrrr@{}}
\toprule
Init. & Upd. &
\shortstack{Block\\NMSE $\downarrow$} &
\shortstack{Block\\Cos. $\uparrow$} &
\shortstack{Output\\NMSE $\downarrow$} &
\shortstack{Output\\Cos. $\uparrow$} \\
\midrule
Random & 1,250 & 0.2886 & 0.8545 & 0.0357 & 0.9820 \\
Manual & 1,250 & 0.2673 & 0.8637 & 0.0320 & 0.9839 \\
Random & 2,500 & 0.2148 & 0.8917 & 0.0258 & 0.9870 \\
Manual & 2,500 & 0.1872 & 0.9049 & 0.0222 & 0.9889 \\
\midrule
\textbf{LinMorph} & 1,250 & \textbf{0.1476} & \textbf{0.9257} & \textbf{0.0235} & \textbf{0.9882} \\
\bottomrule
\end{tabular}
\end{minipage}

\end{table}

\noindent\textbf{Wan2.2: higher block fidelity with half the recovery updates.} The benefit extends to Wan2.2 with six replacements.
Figure~\ref{fig:recovery_curves} shows that LinMorph substantially reduces
the early recovery loss relative to manual copying and random initialization. At the same 1,250-update budget, LinMorph achieves lower block and final-output NMSE than both baselines. Moreover, after only 1,250 updates, LinMorph reaches a block NMSE of $0.14755$, outperforming manual copying and random initialization trained for 2,500 updates, which obtain $0.18718$ and $0.21479$, respectively. Thus, the learned initialization achieves stronger block alignment with half as many recovery updates. Across both backbones, these results demonstrate the effectiveness of learning how pretrained weights should be transferred before recovery.

\subsection{Generation Quality}
\label{sec:generation_quality}

\paragraph{Overall quality across two benchmarks.}
We evaluate the recovered models on VBench~\citep{huang2024vbench}
and VBench-2.0~\citep{zheng2025vbench2}, covering perceptual and
temporal quality as well as intrinsic video faithfulness.
Table~\ref{tab:vbench_totals} consolidates the full-suite total scores
for both conversion depths.
Each LinMorph row uses the same checkpoint across the two benchmarks:
microstep 8,000 for 14 GDN blocks and microstep 40,000 for 21 GDN blocks.

The references include the source backbone, CogVideoX-based generation
and representation-alignment models, and recent Wan-based acceleration
methods.
These provide complementary context for generation quality;
the matched initialization ablations in
Section~\ref{sec:init_ablation} isolate the benefit of learned
weight transfer.

After 4,000 optimizer updates of final full-model fine-tuning,
the 14-GDN model achieves \textbf{82.16} on VBench and
\textbf{55.11} on VBench-2.0.
Its VBench total is within 0.33 points of the reported
Wan2.1-1.3B reference, while both totals exceed the reported
CogVideoX-2B and VideoREPA-2B results.
At a higher conversion ratio of $21/30=70\%$, LinMorph obtains
\textbf{81.85}$^*$ and \textbf{52.13}, respectively,
retaining a VBench total above 81 despite replacing most
attention blocks.

%

\begin{table}[t]
\centering
\footnotesize
\renewcommand{\arraystretch}{0.92}
\setlength{\tabcolsep}{5.0pt}
\caption{\textbf{Overall generation quality on VBench and VBench-2.0.}
Scores are on a 0--100 scale; higher is better.
LinMorph uses 50-step sampling.
$^{\dagger}$ denotes an approximate score read from the
$480\times832$ panel of Figure~5 in
\citet{ghafoorian2025attention}.
Reference provenance and evaluation details are provided in
Appendix~\ref{app:benchmark_details}.}
\label{tab:vbench_totals}

\begin{tabular}{@{}lll@{\hspace{20pt}}lll@{}}
\toprule
Model & VB1 $\uparrow$ & VB2 $\uparrow$
& Model & VB1 $\uparrow$ & VB2 $\uparrow$ \\
\midrule

\multicolumn{6}{c}{\textit{General video-generation models}} \\

Wan2.1-1.3B$^{a}$
& 82.49 & 56.02
&
CogVideoX-1.5-5B$^{c}$
& 82.01 & 53.40 \\

MoAlign-2B$^{b}$
& 81.30 & 55.90
&
CogVideoX-2B$^{b}$
& 80.60 & 54.90 \\

VideoREPA-2B$^{b}$
& 80.50 & 55.00
&
VideoCrafter2$^{d}$
& 80.44 & -- \\

\midrule
\multicolumn{6}{c}{\textit{Efficient video-generation models}} \\

Wan-DMD (2 steps)$^{a}$
& 83.28 & 56.67
&
PyramidalWan-DMD-PT$^{\star,a}$
& 82.72 & 51.75 \\

Wan-Adv (2 steps)$^{a}$
& 82.35 & 54.82
&
SnapGen-V (4 steps)$^{c}$
& 81.14 & -- \\

Wan-Adv (1 step)$^{a}$
& 80.28 & 50.36
&
LTX-Video$^{c}$
& 80.00 & -- \\

Open-Sora V1.2$^{c}$
& 79.76 & --
&
Wan-DMD (1 step)$^{a}$
& 79.45 & 53.17 \\

\midrule
\multicolumn{6}{c}{\textit{Linear-complexity video-generation models}} \\

Attention Surgery(15$\times$R2)$^{e}$
& 83.21 & 55.10
&
M4V$^{c}$
& 81.91 & -- \\

HLA-3F-R1-10$^{e}$
& 81.42 & --
&
Attention Surgery (20$\times$R4)$^{e,\dagger}$
& $\approx 81.1$ & -- \\

\cmidrule(lr){1-6}

\textbf{LinMorph-14}
& 82.16 & 55.11
&
\textbf{LinMorph-21}
& 81.85$^{*}$ & 52.13 \\

\bottomrule
\end{tabular}


\end{table}

\paragraph{Dimension-wise comparison with LinVideo.}
LinVideo~\citep{huang2025linvideo} provides a direct post-training
linearization reference on the same Wan2.1-1.3B backbone.
Table~\ref{tab:linvideo} compares our 21-GDN model with its
16-block conversion using all eight dimensions reported in its
main Wan2.1-1.3B comparison.
This panel jointly covers frame appearance, motion,
temporal consistency, and semantic alignment, allowing us
to examine where quality is retained after conversion.

LinMorph has higher reported scores on
\textbf{seven of eight dimensions}
while converting five additional blocks.
The largest gains are in aesthetic quality ($+4.49$ points),
scene ($+3.32$), and imaging quality ($+2.97$),
accompanied by higher motion smoothness, dynamic degree,
and subject/background consistency.
These results highlight the perceptual and temporal quality
of the recovered 21-GDN model.
Sampling settings and reference sources for both tables are
detailed in Appendix~\ref{app:benchmark_details}.

\begin{table}[t]
\centering
\small
\caption{\textbf{Dimension-wise VBench comparison with LinVideo.}
All eight dimensions in the Wan2.1-1.3B panel of
\citet{huang2025linvideo} are included.
Scores are multiplied by 100; higher is better.
Bold marks the higher score in each row.
LinVideo converts 16/30 blocks; LinMorph converts 21/30.}
\label{tab:linvideo}
\setlength{\tabcolsep}{8pt}
\begin{tabular}{lrrr}
\toprule
Dimension & LinVideo & LinMorph (21 GDN) & $\Delta$ \\
\midrule
Imaging quality
& 66.07 & \textbf{69.04} & $+2.97$ \\
Aesthetic quality
& 59.41 & \textbf{63.90} & $+4.49$ \\
Motion smoothness
& 98.19 & \textbf{98.58} & $+0.39$ \\
Dynamic degree
& 59.67 & \textbf{60.56} & $+0.89$ \\
Background consistency
& 96.72 & \textbf{97.16} & $+0.44$ \\
Subject consistency
& 95.12 & \textbf{95.48} & $+0.36$ \\
Scene
& 39.18 & \textbf{42.50} & $+3.32$ \\
Overall consistency
& \textbf{26.52} & 24.13 & $-2.39$ \\
\bottomrule
\end{tabular}
\par\smallskip
\begin{minipage}{0.98\linewidth}
\footnotesize
$\Delta$: LinMorph minus LinVideo, in score points.
LinVideo uses unaugmented prompts and CFG five;
LinMorph uses Wan-enhanced prompts and CFG six.
This panel excludes temporal flickering and is unaffected by
the prompt correction marked in Table~\ref{tab:vbench_totals}.
\end{minipage}
\end{table}

Qualitative comparisons appear in
Figure~\ref{fig:qualitative_comparison}
(Appendix~\ref{app:qualitative_comparison}).

\subsection{Progressive versus One-Shot Recovery}
\label{sec:progressive_ablation}

Progressive distillation and final joint fine-tuning serve complementary
roles. Distillation adapts the inserted modules while retaining the
pretrained backbone; joint fine-tuning then unfreezes the full DiT to adapt
inter-block representations and the final denoising prediction together.
We first test block alignment at the end of distillation, before this
full-model adaptation.

We compare direct insertion of 21 GDN blocks with
$0\!\rightarrow\!7\!\rightarrow\!14\!\rightarrow\!21$ progressive conversion,
using the same mapped initialization bank and 4,500 total recovery updates.
The progressive schedule uses 1,500 updates per stage and inherits previously
recovered modules; the direct schedule trains all 21 replacements for
4,500 updates.

\begin{table}[t]
\centering
\small
\caption{\textbf{Block alignment after 21-GDN distillation.}
Both schedules use the same mapped initialization bank and 4,500 total
optimizer updates. Metrics use student depth rollout, before final
teacher-free full-model fine-tuning.
Complete prediction diagnostics appear in
Appendix~\ref{app:progressive_diagnostics}.}
\label{tab:progressive}
\begin{tabular}{lrrr}
\toprule
Schedule & Total updates & \shortstack{Block\\NMSE $\downarrow$} & \shortstack{Block\\Cos.\ $\uparrow$} \\
\midrule
One-shot $0\rightarrow21$ & 4,500 & 0.18406 & 0.90358 \\
Progressive $0\rightarrow7\rightarrow14\rightarrow21$ & 4,500 & \textbf{0.17931} & \textbf{0.90728} \\
\bottomrule
\end{tabular}
\end{table}

Progressive recovery reduces depth-rollout block NMSE by $2.58\%$ and
improves block cosine similarity from $0.90358$ to $0.90728$.
This supports progressive insertion as a way to improve intermediate
alignment before joint adaptation.
The final teacher-free stage is designed to accommodate the representation
changes introduced across blocks, allowing retained and replaced components
to adapt together under the generation objective.
Appendix~\ref{app:progressive_diagnostics} reports the complete pre-fine-tuning
diagnostics, including final-prediction metrics.

\subsection{Inference Efficiency}
\label{sec:inference_efficiency}

\begin{wrapfigure}{r}{0.44\textwidth}
    \centering
    \setlength{\abovecaptionskip}{4pt}
    \setlength{\belowcaptionskip}{0pt}
    \includegraphics[width=\linewidth]
    {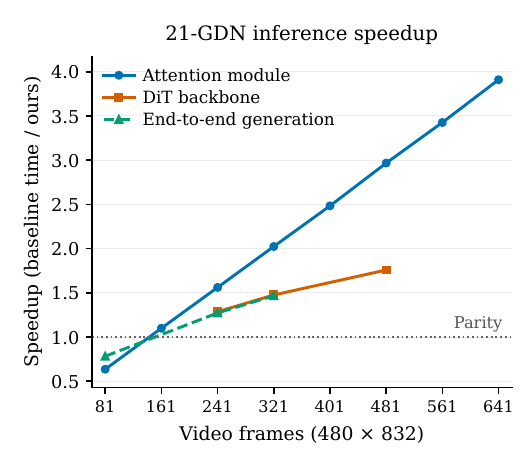}
    \caption{\textbf{21-GDN inference speedup.}
    Attention/backbone: median and IQR over three paired rounds;
    end-to-end: one pair.
    Protocol: Appendix~\ref{app:runtime_details}.}
    \label{fig:runtime_scaling}
    \vspace{-20pt}
\end{wrapfigure}

\paragraph{Larger gains as video length increases.}
We benchmark the 21-GDN hybrid against pretrained Wan2.1 with
FlashAttention-2~\citep{dao2024flashattention2}, using BF16 and
batch size one on an A800-80GB. Figure~\ref{fig:runtime_scaling} reports attention-module, full-backbone, and end-to-end generation speedups using the original FLA implementation. At $480\times832$, attention-module speedup increases from $2.02\times$ at 321 frames to $3.91\times$ at 641 frames. The full DiT reaches $1.48\times$ speedup at 321 frames and $1.76\times$ at 481 frames, showing that the operator-level gains translate into backbone acceleration.

\paragraph{End-to-end acceleration.}
At 321 frames, full generation decreases from $1{,}604.7$ to
$1{,}094.0$ seconds, a $1.47\times$ speedup.
This measurement includes text encoding, denoising, and VAE decoding,
demonstrating a practical reduction in generation latency beyond the
replaced attention modules.
Absolute latencies, resource use, and the measurement protocol are provided
in Appendix~\ref{app:runtime_details}.

\subsection{Transformation Design}
\label{sec:transformation_design}

We compare manual copying and five learned mappings under a common
teacher, target architecture, and evaluation protocol, with equal
fitting budgets for the learned variants.
PreMix/PostMix denotes source-weight mixing before/after dimensional
expansion; residual refinement adds a full-rank correction in native
target space.
Table~\ref{tab:transformation_design} shows that residual refinement
improves every learned variant across all four metrics.
FullMix achieves the best block fidelity, while PostHeadMix achieves
the best output fidelity, both with residual refinement.

\begin{table}[t]
\centering
\footnotesize
\setlength{\tabcolsep}{4pt}
\caption{\textbf{Comparison of weight-transformation designs.}
Entries report results without / with residual refinement.
Block metrics use teacher-forced inputs; output metrics evaluate
the cascade of all 30 mapped modules.
Evaluation uses 64 videos and five diffusion timesteps.
Bold marks the best result in each metric.}
\label{tab:transformation_design}

\begin{tabular}{@{}lrrrr@{}}
\toprule
Transformation
& Block NMSE $\downarrow$
& Block Cos.\ $\uparrow$
& Output NMSE $\downarrow$
& Output Cos.\ $\uparrow$ \\
\midrule
Manual
& 0.1002 / --
& 0.9614 / --
& 4.1482 / --
& 0.2395 / -- \\

FullMix
& 0.0177 / \textbf{0.0161}
& 0.9912 / \textbf{0.9920}
& 0.7810 / 0.6390
& 0.6456 / 0.7051 \\

PreHeadMix
& 0.0221 / 0.0207
& 0.9890 / 0.9897
& 1.5084 / 1.2746
& 0.5176 / 0.5745 \\

PostHeadMix
& 0.0175 / 0.0162
& 0.9913 / 0.9919
& 0.7376 / \textbf{0.6166}
& 0.6625 / \textbf{0.7135} \\

PreChannelMix
& 0.0183 / 0.0166
& 0.9908 / 0.9917
& 0.9758 / 0.6926
& 0.6071 / 0.6893 \\

PostChannelMix
& 0.0207 / 0.0181
& 0.9897 / 0.9910
& 1.4928 / 0.8154
& 0.4759 / 0.6536 \\
\bottomrule
\end{tabular}
\end{table}

\section{Conclusion}
We presented LinMorph, a framework that treats post-hoc Video DiT
linearization as a learnable weight transformation problem rather
than solely a recovery problem. Instead of random initialization or
rigid manual copying, LinMorph learns cross-parameter correspondences
that transfer pretrained softmax knowledge into GDN parameters,
including those without direct source counterparts. A single
source-conditioned morphism is shared across layers while producing
layer-specific initializations. Operating in weight space, it can
be fitted on short, low-resolution clips and reused at larger video
scales without refitting. Experiments on two Wan backbones
demonstrate improved initialization fidelity and faster recovery,
while the resulting hybrid models achieve strong generation quality
and accelerate long-video inference. These findings establish
learned initialization as a complementary axis for efficient
linearization: how pretrained knowledge is transferred before
recovery matters alongside how the converted model is subsequently
optimized.

\subsection*{AI use statement}

Generative AI tools were used to assist with literature discovery and
synthesis, methodological and experimental feedback, software and plotting
support, interpretation and presentation of experimental results, and
manuscript organization, translation, and language editing.
All AI-assisted code, analyses, citations, claims, and text are subject to
author review and verification against the implementation, experimental
records, and original sources. The authors take responsibility for the
final content of this work, including material produced with the aid of
generative AI.

\subsection*{Ethics statement}

This work studies the efficient adaptation and inference of pretrained
video diffusion models. The resulting generators may inherit biases and
misuse risks from their pretrained backbones, including the creation of
misleading or non-consensual media. Responsible deployment should include
appropriate content safeguards and clear identification of synthetic
media. Our experiments use pretrained models, existing datasets, and a
curated set of high-quality internet videos. The use and any redistribution
of these resources should respect applicable licenses, copyright, privacy,
and consent requirements.

\subsection*{Reproducibility statement}

Section~\ref{sec:linmorph} describes the shared weight morphism and its
training objectives. Appendix~\ref{app:linmorph} specifies target-parameter
construction, dimensional alignment, initialization, optimization, and
progressive replacement.
Appendix~\ref{app:experimental_protocol} documents backbone-specific
recovery settings, baseline initialization recipes, and checkpoint
accounting. Functional evaluation metrics and statistical scope are
provided in Appendix~\ref{app:evaluation_protocol}; generation settings
and benchmark reference sources are detailed in
Appendix~\ref{app:benchmark_details}.
Appendix~\ref{app:runtime_details} reports the timing environment,
measurement scopes, warmup procedures, and aggregation protocol.

\bibliographystyle{unsrtnat}
\bibliography{reference}

@inproceedings{peebles2023scalable,
  title     = {Scalable Diffusion Models with Transformers},
  author    = {Peebles, William and Xie, Saining},
  booktitle = {Proceedings of the IEEE/CVF International Conference on Computer Vision},
  pages     = {4195--4205},
  year      = {2023}
}

@article{ma2024latte,
    title={Latte: Latent Diffusion Transformer for Video Generation},
    author={Xin Ma and Yaohui Wang and Xinyuan Chen and Gengyun Jia and Ziwei Liu and Yuan-Fang Li and Cunjian Chen and Yu Qiao},
    journal={Transactions on Machine Learning Research},
    issn={2835-8856},
    year={2025},
    url={https://openreview.net/forum?id=ntGPYNUF3t},
    note={}
}

@inproceedings{yang2024cogvideox,
    title={CogVideoX: Text-to-Video Diffusion Models with An Expert Transformer},
    author={Zhuoyi Yang and Jiayan Teng and Wendi Zheng and Ming Ding and Shiyu Huang and Jiazheng Xu and Yuanming Yang and Wenyi Hong and Xiaohan Zhang and Guanyu Feng and Da Yin and Yuxuan Zhang and Weihan Wang and Yean Cheng and Bin Xu and Xiaotao Gu and Yuxiao Dong and Jie Tang},
    booktitle={The Thirteenth International Conference on Learning Representations},
    year={2025},
    url={https://openreview.net/forum?id=LQzN6TRFg9}
}

@article{kong2024hunyuanvideo,
  title   = {{HunyuanVideo}: A Systematic Framework for Large Video Generative Models},
  author  = {Kong, Weijie and Tian, Qi and Zhang, Zijian and Min, Rox and Dai, Zuozhuo and Zhou, Jin and Xiong, Jiangfeng and Li, Xin and Wu, Bo and Zhang, Jianwei and others},
  journal = {arXiv preprint arXiv:2412.03603},
  year    = {2024}
}

@article{wan2025,
  title   = {Wan: Open and Advanced Large-Scale Video Generative Models},
  author  = {{Team Wan}},
  journal = {arXiv preprint arXiv:2503.20314},
  year    = {2025}
}

@inproceedings{katharopoulos2020transformers,
  title     = {Transformers are {RNNs}: Fast Autoregressive Transformers with Linear Attention},
  author    = {Katharopoulos, Angelos and Vyas, Apoorv and Pappas, Nikolaos and Fleuret, Fran{\c{c}}ois},
  booktitle = {Proceedings of the 37th International Conference on Machine Learning},
  series    = {Proceedings of Machine Learning Research},
  volume    = {119},
  pages     = {5156--5165},
  year      = {2020}
}

@inproceedings{yang2024gla,
  title     = {Gated Linear Attention Transformers with Hardware-Efficient Training},
  author    = {Yang, Songlin and Wang, Bailin and Shen, Yikang and Panda, Rameswar and Kim, Yoon},
  booktitle = {Proceedings of the 41st International Conference on Machine Learning},
  series    = {Proceedings of Machine Learning Research},
  volume    = {235},
  pages     = {56501--56523},
  year      = {2024}
}

@inproceedings{schlag2021linear,
  title     = {Linear Transformers Are Secretly Fast Weight Programmers},
  author    = {Schlag, Imanol and Irie, Kazuki and Schmidhuber, J{\"u}rgen},
  booktitle = {Proceedings of the 38th International Conference on Machine Learning},
  series    = {Proceedings of Machine Learning Research},
  volume    = {139},
  pages     = {9355--9366},
  year      = {2021}
}

@inproceedings{yang2024deltanet,
  title     = {Parallelizing Linear Transformers with the Delta Rule over Sequence Length},
  author    = {Yang, Songlin and Wang, Bailin and Zhang, Yu and Shen, Yikang and Kim, Yoon},
  booktitle = {Advances in Neural Information Processing Systems},
  year      = {2024}
}

@inproceedings{yang2024gateddeltanet,
    title={Gated Delta Networks: Improving Mamba2 with Delta Rule},
    author={Songlin Yang and Jan Kautz and Ali Hatamizadeh},
    booktitle={The Thirteenth International Conference on Learning Representations},
    year={2025},
    url={https://openreview.net/forum?id=r8H7xhYPwz}
}

@inproceedings{wang2025lingen,
  title     = {{LinGen}: Towards High-Resolution Minute-Length Text-to-Video Generation with Linear Computational Complexity},
  author    = {Wang, Hongjie and Ma, Chih-Yao and Liu, Yen-Cheng and Hou, Ji and Xu, Tao and Wang, Jialiang and Juefei-Xu, Felix and Luo, Yaqiao and Zhang, Peizhao and Hou, Tingbo and Vajda, Peter and Jha, Niraj K. and Dai, Xiaoliang},
  booktitle = {Proceedings of the IEEE/CVF Conference on Computer Vision and Pattern Recognition},
  year      = {2025}
}

@inproceedings{chen2025sanavideo,
    title={{SANA}-Video: Efficient Video Generation with Block Linear Diffusion Transformer},
    author={Junsong Chen and Yuyang Zhao and Jincheng YU and Ruihang Chu and Junyu Chen and Shuai Yang and Xianbang Wang and Yicheng Pan and Daquan Zhou and Huan Ling and Haozhe Liu and Hongwei Yi and Hao Zhang and Muyang Li and Yukang Chen and Han Cai and Sanja Fidler and Ping Luo and Song Han and Enze Xie},
    booktitle={The Fourteenth International Conference on Learning Representations},
    year={2026},
    url={https://openreview.net/forum?id=mzAchylAtf}
}

@inproceedings{liu2024clear,
  title     = {{CLEAR}: Conv-Like Linearization Revs Pre-Trained Diffusion Transformers Up},
  author    = {Liu, Songhua and Tan, Zhenxiong and Wang, Xinchao},
  booktitle = {Advances in Neural Information Processing Systems},
  year      = {2025},
  eprint    = {2412.16112},
  archiveprefix = {arXiv},
  url       = {https://papers.nips.cc/paper_files/paper/2025/hash/e36da7acd188c6655792270b38830124-Abstract-Conference.html}
}

@inproceedings{becker2025edit,
  title     = {{EDiT}: Efficient Diffusion Transformers with Linear Compressed Attention},
  author    = {Becker, Philipp and Mehrotra, Abhinav and Chavhan, Ruchika and Chadwick, Malcolm and Morreale, Luca and Noroozi, Mehdi and Gil C. P. Ramos, Alberto and Bhattacharya, Sourav},
  booktitle = {Proceedings of the IEEE/CVF International Conference on Computer Vision},
  pages     = {19608--19616},
  year      = {2025}
}

@inproceedings{wang2025lit,
  title     = {{LiT}: Delving into a Simple Linear Diffusion Transformer for Image Generation},
  author    = {Wang, Jiahao and Kang, Ning and Yao, Lewei and Chen, Mengzhao and Wu, Chengyue and Zhang, Songyang and Xue, Shuchen and Liu, Yong and Wu, Taiqiang and Liu, Xihui and Zhang, Kaipeng and Zhang, Shifeng and Shao, Wenqi and Li, Zhenguo and Luo, Ping},
  booktitle = {Proceedings of the IEEE/CVF International Conference on Computer Vision},
  pages     = {16068--16078},
  year      = {2025}
}

@inproceedings{ghafoorian2025attention,
  title     = {Attention Surgery: An Efficient Recipe to Linearize Your Video Diffusion Transformer},
  author    = {Ghafoorian, Mohsen and Korzhenkov, Denis and Habibian, Amirhossein},
  booktitle = {Proceedings of the IEEE/CVF Conference on Computer Vision and Pattern Recognition},
  pages     = {32915--32925},
  year      = {2026}
}

@inproceedings{huang2025linvideo,
  title     = {{LinVideo}: A Post-Training Framework towards {$O(n)$} Attention in Efficient Video Generation},
  author    = {Huang, Yushi and Ge, Xingtong and Gong, Ruihao and Lv, Chengtao and Zhang, Jun},
  booktitle = {Proceedings of the IEEE/CVF Conference on Computer Vision and Pattern Recognition},
  pages     = {23398--23408},
  year      = {2026}
}

@inproceedings{ghafoorian2026rehyat,
  title     = {{ReHyAt}: Recurrent Hybrid Attention for Video Diffusion Transformers},
  author    = {Ghafoorian, Mohsen and Habibian, Amirhossein},
  booktitle = {Proceedings of the IEEE/CVF Conference on Computer Vision and Pattern Recognition},
  pages     = {40674--40684},
  year      = {2026}
}

@inproceedings{kasai2021finetuning,
  title     = {Finetuning Pretrained Transformers into {RNN}s},
  author    = {Kasai, Jungo and Peng, Hao and Zhang, Yizhe and Yogatama, Dani and
  Ilharco, Gabriel and Pappas, Nikolaos and Mao, Yi and Chen, Weizhu and Smith, Noah A.},
  booktitle = {Proceedings of the 2021 Conference on Empirical Methods in Natural Language Processing},
  pages     = {10630--10643},
  year      = {2021},
  doi       = {10.18653/v1/2021.emnlp-main.830},
  url       = {https://aclanthology.org/2021.emnlp-main.830/}
}

@inproceedings{chen2015net2net,
  title     = {{Net2Net}: Accelerating Learning via Knowledge Transfer},
  author    = {Chen, Tianqi and Goodfellow, Ian J. and Shlens, Jonathon},
  booktitle = {4th International Conference on Learning Representations, {ICLR} 2016,
               San Juan, Puerto Rico, May 2--4, 2016, Conference Track Proceedings},
  editor    = {Bengio, Yoshua and LeCun, Yann},
  year      = {2016},
  eprint    = {1511.05641},
  archiveprefix = {arXiv},
  url       = {https://arxiv.org/abs/1511.05641}
}

@inproceedings{wei2016networkmorphism,
  title     = {Network Morphism},
  author    = {Wei, Tao and Wang, Changhu and Rui, Yong and Chen, Chang Wen},
  booktitle = {Proceedings of the 33rd International Conference on Machine Learning},
  series    = {Proceedings of Machine Learning Research},
  volume    = {48},
  pages     = {564--572},
  year      = {2016}
}

@inproceedings{he2025joint,
  title     = {Joint Fine-tuning and Conversion of Pretrained Speech and Language Models towards Linear Complexity},
  author    = {He, Mutian and Garner, Philip N.},
  booktitle = {International Conference on Learning Representations},
  year      = {2025}
}

@inproceedings{feng2025wave,
  title     = {{WAVE}: Weight Templates for Adaptive Initialization of Variable-Sized Models},
  author    = {Feng, Fu and Xie, Yucheng and Wang, Jing and Geng, Xin},
  booktitle = {Proceedings of the IEEE/CVF Conference on Computer Vision and Pattern Recognition},
  pages     = {4819--4828},
  year      = {2025}
}

@inproceedings{xu2026pith,
  title     = {Unlocking Pre-Trained Weights: Parameter Inheritance for Zero-Shot Initialization},
  author    = {Xu, Jiaze and Xia, Shiyu and Lv, Jiaqi and Geng, Xin},
  booktitle = {Proceedings of the IEEE/CVF Conference on Computer Vision and Pattern Recognition},
  pages     = {34502--34511},
  year      = {2026}
}

@inproceedings{huang2024vbench,
  title = {{VBench}: Comprehensive Benchmark Suite for Video Generative Models},
  author = {Huang, Ziqi and He, Yinan and Yu, Jiashuo and Zhang, Fan and Si, Chenyang and Jiang, Yuming and Zhang, Yuanhan and Wu, Tianxing and Jin, Qingyang and Chanpaisit, Nattapol and Wang, Yaohui and Chen, Xinyuan and Wang, Limin and Lin, Dahua and Qiao, Yu and Liu, Ziwei},
  booktitle = {Proceedings of the IEEE/CVF Conference on Computer Vision and Pattern Recognition},
  pages = {21807--21818},
  year = {2024},
  url = {https://openaccess.thecvf.com/content/CVPR2024/html/Huang_VBench_Comprehensive_Benchmark_Suite_for_Video_Generative_Models_CVPR_2024_paper.html}
}

@article{zheng2025vbench2,
  title = {{VBench-2.0}: Advancing Video Generation Benchmark Suite for Intrinsic Faithfulness},
  author = {Zheng, Dian and Huang, Ziqi and Liu, Hongbo and Zou, Kai and He, Yinan and Zhang, Fan and Gu, Lulu and Zhang, Yuanhan and He, Jingwen and Zheng, Wei-Shi and Qiao, Yu and Liu, Ziwei},
  journal = {arXiv preprint arXiv:2503.21755},
  year = {2025},
  url = {https://arxiv.org/abs/2503.21755}
}

@inproceedings{bhowmik2026moalign,
  title = {{MoAlign}: Motion-Centric Representation Alignment for Video Diffusion Models},
  author = {Bhowmik, Aritra and Korzhenkov, Denis and Snoek, Cees G. M. and Habibian, Amirhossein and Ghafoorian, Mohsen},
  booktitle = {International Conference on Learning Representations},
  year = {2026},
  url = {https://openreview.net/forum?id=OR0ySm4l9h}
}

@inproceedings{korzhenkov2026pyramidalwan,
  title = {{PyramidalWan}: On Making Pretrained Video Model Pyramidal for Efficient Inference},
  author = {Korzhenkov, Denis and Karjauv, Adil and Karnewar, Animesh and Ghafoorian, Mohsen and Habibian, Amirhossein},
  booktitle = {Proceedings of the IEEE/CVF Conference on Computer Vision and Pattern Recognition},
  year = {2026},
  url = {https://openaccess.thecvf.com/content/CVPR2026/papers/Korzhenkov_PyramidalWan_On_Making_Pretrained_Video_Model_Pyramidal_for_Efficient_Inference_CVPR_2026_paper.pdf}
}

@inproceedings{chen2024videocrafter2,
  title     = {{VideoCrafter2}: Overcoming Data Limitations for
               High-Quality Video Diffusion Models},
  author    = {Chen, Haoxin and Zhang, Yong and Cun, Xiaodong and
               Xia, Menghan and Wang, Xintao and Weng, Chao and Shan, Ying},
  booktitle = {Proceedings of the IEEE/CVF Conference on Computer
               Vision and Pattern Recognition},
  pages     = {7310--7320},
  year      = {2024},
  url       = {https://openaccess.thecvf.com/content/CVPR2024/html/Chen_VideoCrafter2_Overcoming_Data_Limitations_for_High-Quality_Video_Diffusion_Models_CVPR_2024_paper.html}
}

@article{zheng2024opensora,
  title   = {{Open-Sora}: Democratizing Efficient Video Production for All},
  author  = {Zheng, Zangwei and Peng, Xiangyu and Yang, Tianji and
             Shen, Chenhui and Li, Shenggui and Liu, Hongxin and
             Zhou, Yukun and Li, Tianyi and You, Yang},
  journal = {arXiv preprint arXiv:2412.20404},
  year    = {2024},
  url     = {https://arxiv.org/abs/2412.20404}
}

@article{hacohen2024ltxvideo,
  title   = {{LTX-Video}: Realtime Video Latent Diffusion},
  author  = {HaCohen, Yoav and Chiprut, Nisan and Brazowski, Benny and
             Shalem, Daniel and Moshe, Dudu and Richardson, Eitan and
             Levin, Eran and Shiran, Guy and Zabari, Nir and Gordon, Ori and
             Panet, Poriya and Weissbuch, Sapir and Kulikov, Victor and
             Bitterman, Yaki and Melumian, Zeev and Bibi, Ofir},
  journal = {arXiv preprint arXiv:2501.00103},
  year    = {2024},
  url     = {https://arxiv.org/abs/2501.00103}
}

@inproceedings{wu2025snapgenv,
  title     = {{SnapGen-V}: Generating a Five-Second Video Within
               Five Seconds on a Mobile Device},
  author    = {Wu, Yushu and Zhang, Zhixing and Li, Yanyu and Xu, Yanwu and
               Kag, Anil and Sui, Yang and Coskun, Huseyin and Ma, Ke and
               Lebedev, Aleksei and Hu, Ju and Metaxas, Dimitris N. and
               Wang, Yanzhi and Tulyakov, Sergey and Ren, Jian},
  booktitle = {Proceedings of the IEEE/CVF Conference on Computer
               Vision and Pattern Recognition},
  pages     = {2479--2490},
  year      = {2025},
  doi       = {10.1109/CVPR52734.2025.00237}
}

@inproceedings{huang2026m4v,
  title     = {{M4V}: Multimodal Mamba for Efficient
               Text-to-Video Generation},
  author    = {Huang, Jiancheng and Zhang, Gengwei and Jie, Zequn and
               Jiao, Siyu and Qian, Yinlong and Chen, Ling and
               Wei, Yunchao and Ma, Lin},
  booktitle = {Proceedings of the IEEE/CVF Conference on Computer
               Vision and Pattern Recognition},
  pages     = {36747--36757},
  year      = {2026},
  url       = {https://openaccess.thecvf.com/content/CVPR2026/html/Huang_M4V_Multimodal_Mamba_for_Efficient_Text-to-Video_Generation_CVPR_2026_paper.html}
}

@article{ackermann2026hla,
  title         = {{HLA}: Hadamard Linear Attention},
  author        = {Ackermann, Hanno and Cai, Hong and Ghafoorian, Mohsen and
                   Habibian, Amirhossein},
  journal       = {arXiv preprint arXiv:2602.12128},
  year          = {2026},
  eprint        = {2602.12128},
  archiveprefix = {arXiv},
  url           = {https://arxiv.org/abs/2602.12128}
}

@inproceedings{dao2024flashattention2,
  title = {{FlashAttention-2}: Faster Attention with Better Parallelism and Work Partitioning},
  author = {Dao, Tri},
  booktitle = {The Twelfth International Conference on Learning Representations},
  year = {2024},
  url = {https://arxiv.org/abs/2307.08691}
}

@misc{wan2025wan22,
  author       = {{Team Wan}},
  title        = {{Wan2.2}: Official Models and Code},
  year         = {2025},
  howpublished = {GitHub repository},
  url          = {https://github.com/Wan-Video/Wan2.2}
}

@inproceedings{nan2025openvid,
  author    = {Nan, Kepan and Xie, Rui and Zhou, Penghao and
               Fan, Tiehan and Yang, Zhenheng and Chen, Zhijie and
               Li, Xiang and Yang, Jian and Tai, Ying},
  title     = {{OpenVid-1M}: A Large-Scale High-Quality Dataset
               for Text-to-video Generation},
  booktitle = {The Thirteenth International Conference on
               Learning Representations},
  year      = {2025},
  url       = {https://openreview.net/forum?id=j7kdXSrISM}
}

\appendix

\section{Related Work}

\paragraph{Video diffusion transformers.}
Diffusion Transformers (DiTs) replace the convolutional backbone of diffusion
models with Transformer blocks and exhibit favorable scaling behavior for
visual generation~\citep{peebles2023scalable}.
This paradigm has been extended to video generation by modeling videos as
sequences of spatio-temporal latent tokens.
Early works such as Latte investigate factorized spatial--temporal Transformer
designs~\citep{ma2024latte}, while recent large-scale systems, including
CogVideoX~\citep{yang2024cogvideox}, HunyuanVideo~\citep{kong2024hunyuanvideo},
and Wan~\citep{wan2025}, demonstrate the strong generation quality enabled by
scaling Video DiTs.
However, increasing the spatial resolution or video duration rapidly enlarges
the token sequence, making quadratic self-attention a major computational and
memory bottleneck.
Our work builds on pretrained Video DiTs and focuses on reducing this cost
without repeating their expensive pretraining.

\paragraph{Efficient attention and Video DiT linearization.}
Linear attention reduces the quadratic token-interaction cost by rearranging
kernelized attention computation~\citep{katharopoulos2020transformers}.
Subsequent recurrent formulations improve its expressiveness through
data-dependent gating~\citep{yang2024gla}, delta-rule memory
updates~\citep{schlag2021linear,yang2024deltanet}, or their
combination~\citep{yang2024gateddeltanet}.
These operators introduce state-update, decay, and gating parameters that do
not have direct counterparts in conventional softmax attention.

Efficient attention has also been explored in diffusion models.
LinGen~\citep{wang2025lingen} and SANA-Video~\citep{chen2025sanavideo}
construct video generators around linear-complexity backbones, but require
dedicated model training.
For pretrained image DiTs, CLEAR~\citep{liu2024clear},
EDiT~\citep{becker2025edit}, and LiT~\citep{wang2025lit} study architectural
replacement and distillation recipes for efficient attention.
More closely related to our setting, Attention
Surgery~\citep{ghafoorian2025attention}, LinVideo~\citep{huang2025linvideo},
and ReHyAt~\citep{ghafoorian2026rehyat} convert pretrained Video DiTs through
hybrid attention, selective layer replacement, attention distillation, and
subsequent recovery.
These methods primarily improve \emph{which} attention operator or layer to
replace and \emph{how} the converted model should be recovered.
LinMorph is complementary: it addresses \emph{how the parameters of the
replacement operator should be initialized} before recovery.

\paragraph{Cross-architecture parameter transfer.}
A related line of research transfers knowledge between neural networks through
their parameters.
Net2Net~\citep{chen2015net2net} and network
morphism~\citep{wei2016networkmorphism} construct function-preserving
transformations when changing network depth, width, or kernel size.
Transformer-to-RNN conversion instead replaces pretrained attention with a
recurrent alternative and recovers the converted model through
fine-tuning~\citep{kasai2021finetuning}, while cross-architecture layer-wise
distillation transfers intermediate behavior between structurally different
models~\citep{he2025joint}.
Recent parameter-prediction methods further learn weight templates or
projections for initializing networks of different
configurations~\citep{feng2025wave,xu2026pith}.
These lines of work study function-preserving network changes, recurrent
conversion, cross-architecture distillation, and initialization across model
configurations.
Our focus is a shared, source-conditioned initialization rule for
softmax-to-GDN conversion. It uses pretrained projections to initialize both
corresponding parameters and unmatched components such as gates, decays, and
update projections, before recovering the converted model.

\section{Implementation Details of LinMorph}
\label{app:linmorph}

\subsection{Notation and Parameter Conventions}
\label{app:linmorph_notation}

All projection weights use the output-by-input convention:
$\mathbf{W}\in\mathbb{R}^{d_{\mathrm{out}}\times d_{\mathrm{in}}}$ acts as
$\mathbf{X}\mathbf{W}^{\top}$ on row-token inputs.
Uppercase source subscripts $Q,K,V,O$ distinguish pretrained projections
from lowercase target subscripts $q,k,v,o,g,a,\beta$.
The subscript $T$ identifies teacher activations, while matrix
transposition is denoted by $\top$.
An overbar denotes a source-conditioned base weight before residual
correction, and a tilde denotes the resulting initialization after
residual correction.
Hats denote outputs of the mapped replacement during local alignment.
The subscript $\mathrm{mix}$ denotes the source mixture before
hidden-width alignment.
These distinctions apply during mapper fitting; after materialization,
the target parameters are independently optimized and need not remain
representable by the original shared mapper.

\begin{table}[t]
    \centering
    \small
    \caption{Notation used throughout the manuscript.}
    \label{tab:linmorph_notation}
    \begin{tabular}{lp{0.64\linewidth}}
        \toprule
        Symbol & Meaning\\
        \midrule
        $\ell$ & Transformer block index.\\
        $s,\mathcal{S},S$ & Source index, source set $\{Q,K,V,O\}$, and its size $S=4$.\\
        $j,b,i$ & Target-parameter index, mixing-group index, and channel index.\\
        $t,\sigma_t,c$ & Diffusion scheduler timestep, corresponding noise level, and text condition.\\
        $D,H_s,d_s$ & Teacher width, head count, and head width, with $D=H_sd_s$.\\
        $D_g,H,d_k,d_v$ & GDN hidden width, recurrent head count, and per-head key/value widths.\\
        $C_j,G_j$ & Attention-channel count and mixing-group count for target $j$.\\
        $\alpha_{j,b,s}$ & Scalar coefficient for source $s$ in target group $(j,b)$.\\
        $\boldsymbol{\alpha}_{j,b}$ & The corresponding $S$-dimensional coefficient vector.\\
        $\phi,\psi$ & Shared base-mapping parameters and shared residual parameters.\\
        $\widetilde{\Theta}_{\mathrm{GDN}}^{(\ell)}$ & Generated native GDN initialization, excluding hidden-width bridges.\\
        $\widetilde{}{\Theta}_{\mathrm{lin}}^{(\ell)}$ & Generated complete replacement parameters, including both bridges.\\
        $\Theta_{\mathrm{lin}}^{(\ell,0)}$ & Materialized initialization produced by the fitted mapper.\\
        $\mathcal{B},\mathcal{R}_m$ & Layers used for local mapping loss and layers replaced at recovery stage $m$.\\
        $m,m_\star$ & Progressive replacement stage and chosen final replacement stage ($2$ or $3$).\\
        $\theta_m,\theta$ & Student model parameters during progressive stage $m$ and final joint recovery.\\
        $v_T,v_{\theta_m}$ & Frozen teacher and stage-$m$ student velocity predictions.\\
        \bottomrule
    \end{tabular}
\end{table}

\subsection{Target Operator, Shapes, and Initialization}
\label{app:linmorph_shapes}

The full-model configuration uses a 30-block Wan2.1-T2V-1.3B teacher with
$D=1536$, $H_s=12$ source heads, and source head width $d_s=128$.
The target GDN uses $D_g=3072$, $H=24$, $d_k=256$, and $d_v=512$.
Consequently, the query/key channel count is $C_k=6144$ and the value/gate
channel count is $C_v=12288$; neither should be confused with the internal
hidden width $D_g$.
Each replacement performs a forward causal scan over the flattened video-token
sequence, with short convolutions, $\ell_2$-normalized queries and keys,
sigmoid update gates, learned decay, and gated output normalization.
The state is reset for each attention evaluation.
Only the teacher's four projection matrices supply source weights:
teacher projection biases, query/key normalization, and rotary embeddings
are not transferred by the mapping. The external block interface,
cross-attention, feed-forward network, and residual/modulation structure are
retained.

The $q,k,v,o$ mixtures initially select their namesake sources; the $g$ and
$\beta$ mixtures and the auxiliary offsets initially vanish.
Dense channel adapters start with head-aware repetition for $q,k,v,g$ and
corresponding averaging for $o$, and are subsequently learned without
structural constraints. More explicitly, target head $h$ initially draws from
source head $h_s(h)=1+\lfloor(h-1)/r_H\rfloor$, where $r_H=H/H_s$
and $h=1,\ldots,H$.
Its $d_s$-wide source channels are repeated $d_k/d_s$ or $d_v/d_s$ times.
The output adapter averages the $r_H(d_v/d_s)$ repeated copies.

For $r_D=D_g/D$, the hidden bridges initially satisfy
\begin{equation}
    \mathbf{B}_{\mathrm{up}}
    =\begin{bmatrix}\mathbf{I}_D\\ \vdots\\ \mathbf{I}_D\end{bmatrix},
    \qquad
    \mathbf{B}_{\mathrm{down}}
    =\frac{1}{r_D}\mathbf{B}_{\mathrm{up}}^\top,
    \qquad
    \mathbf{B}_{\mathrm{down}}\mathbf{B}_{\mathrm{up}}=\mathbf{I}_D.
\end{equation}
The identity relation describes their initialization only; both bridges
are learned. It does not imply equivalence between softmax and GDN.

\paragraph{Mixing order and group counts.}
The verified full-model recipe expands each source before mixing.
For FullMix, a source-shared linear adapter commutes with the scalar source
mixture, e.g.,
$\sum_s\alpha_s\mathbf{E}\mathbf{W}_s
=\mathbf{E}\sum_s\alpha_s\mathbf{W}_s$.
This equivalence does not generally extend to group-dependent coefficients.
For the main projections, post-expansion HeadMix has $H(d_k/d_s)=48$ groups
for each of $q,k$ and $H(d_v/d_s)=96$ groups for each of $v,g,o$.
With $S=4$ sources, a target with $C_j$ attention channels and $G_j$ groups uses
$S$, $SG_j$, or $SC_j$ coefficients in FullMix, HeadMix, or ChannelMix,
respectively. The schematic therefore uses mixing-group labels rather than identifying
$G_j$ with the $H=24$ recurrent heads.

\subsection{Auxiliary Parameter Construction}
\label{app:linmorph_auxiliary}

\paragraph{Inventory of transformed parameters.}
LinMorph constructs the query, key, value, and output projection
weights $(\mathbf{W}_q,\mathbf{W}_k,\mathbf{W}_v,\mathbf{W}_o)$,
the output-gate projection $\mathbf{W}_g$,
the update-gate projection $\mathbf{W}_{\beta}$,
and the decay projection $\mathbf{W}_a$.
It also constructs the per-head decay parameters $A_{\log}$
and $\mathrm{dt}$ bias, the three short-convolution kernels
applied to the query, key, and value streams, and the learned
scale of the output normalization.
The hidden-width alignment bridges are handled separately
from these native GDN parameters.

The projections $q,k,v,o,g$ are generated through the
shape-aligned matrix mixing described in the main text.
Although $g$ has no namesake softmax counterpart, its projection
is constructed from the same four pretrained source matrices.
The update and decay projections use head-pooled source weights,
while the remaining parameters use reduced source-weight
statistics.
We detail these constructions below and summarize their native
shapes, source reductions, and fixed bases in
Table~\ref{tab:linmorph_auxiliary}.

We omit layer superscripts here. Reshape each stored source projection as
$\mathbf{W}_s[h,u,i]$, with source head $h$, within-head channel $u$, and input
channel $i$. Define the row-pooled matrix and three RMS profiles:
\begin{equation}
\begin{aligned}
    U_s[h,i]&=\frac{1}{d_s}\sum_u W_s[h,u,i],\\
    r_s^{\mathrm{head}}[h]
        &=\operatorname{std}_h\!\left(
            \sqrt{\frac{1}{d_sD}\sum_{u,i}W_s[h,u,i]^2}\right),\\
    r_s^{\mathrm{channel}}[h,u]
        &=\operatorname{std}_{h,u}\!\left(
            \sqrt{\frac{1}{D}\sum_iW_s[h,u,i]^2}\right),\\
    r_s^{\mathrm{norm}}[u]
        &=\operatorname{std}_u\!\left(
            \sqrt{\frac{1}{H_sD}\sum_{h,i}W_s[h,u,i]^2}\right).
\end{aligned}
\label{eq:linmorph_aux_stats}
\end{equation}
Here $\operatorname{std}_{\mathcal{I}}$ denotes centering and division by the
centered RMS over index set $\mathcal{I}$, not the sample standard deviation:
\begin{equation}
    \operatorname{std}_{\mathcal{I}}(x)
    =\frac{x-\operatorname{mean}_{\mathcal{I}}x}
           {\max\!\left(
             \sqrt{\operatorname{mean}_{\mathcal{I}}
                 (x-\operatorname{mean}_{\mathcal{I}}x)^2},
             10^{-6}\right)}.
\end{equation}
Each source is standardized independently. $U_s$ is a plain arithmetic mean
and is not standardized.

In FullMix, each auxiliary target has its own four-source coefficient vector.
The construction for every target consists of reduction, source mixing,
deterministic head/channel repetition, and addition to a fixed base.
Head-indexed quantities use the source-head assignment
$h_s(h)=1+\lfloor(h-1)/r_H\rfloor$.
Within-head channel profiles are repeated to the target head width.
Table~\ref{tab:linmorph_auxiliary} gives the target-specific rules.

\begin{table}[t]
    \centering
    \small
    \caption{Auxiliary mappings in FullMix. Shapes use stored-weight
    conventions. The convolution channel count $C_j$ is $C_k$ for $q,k$
    and $C_v$ for $v$; $K$ is the short-convolution kernel length.}
    \label{tab:linmorph_auxiliary}
    \begin{tabular}{llll}
        \hline
        Target & Native shape & Source reduction & Fixed base\\
        \hline
        Decay projection $a$ & $H\times D_g$ & $U_s$ & Zero\\
        Update projection $\beta$ & $H\times D_g$ & $U_s$ & Zero\\
        $A_{\log}$, $\mathrm{dt}$ bias & $H$ & $r_s^{\mathrm{head}}$ & Native initialization\\
        Short-conv weights & $C_j\times1\times K$ &
            $r_s^{\mathrm{channel}}$ & Identity kernel\\
        Output-norm scale & $d_v$ & $r_s^{\mathrm{norm}}$ & Ones\\
        \hline
    \end{tabular}
\end{table}

For example, FullMix constructs the $a$ and $\beta$ projection bases as
\begin{equation}
    \overline{\mathbf{W}}_j[h,:]
    =\left(\sum_{s\in\mathcal{S}}\alpha_{j,s}
        U_s[h_s(h),:]\right)\mathbf{B}_{\mathrm{down}},
    \qquad j\in\{a,\beta\}.
\end{equation}
Both have shape $H\times D_g$ and zero fixed base.
Here $\overline{\mathbf{W}}_a$ is the source-conditioned decay projection.
The $\beta$ projection additionally receives the native residual in
Equation~\ref{eq:linmorph_residual}; $a$ does not.
For FullMix convolutions, the mixed channel profile is added to every kernel
tap. The base kernel has a one at the last causal tap and zeros elsewhere.
This describes the kernel initialization, not an identity of the entire
nonlinear convolution module. The short convolutions use kernel size $K=4$, no bias, and SiLU activation.

\paragraph{Finer-grained auxiliary variants.}
Auxiliary axes do not always admit the same grouping as a main projection.
For HeadMix, $a$, $A_{\log}$, and the $\mathrm{dt}$ bias use a global mixture
plus a per-target-head mixture. Convolutions use per-tap global mixtures plus
per-tap, per-target-head mixtures. The $\beta$ projection has a per-target-head
mixture, whereas the output-norm scale has only a global mixture because it
has no separate head axis.
For ChannelMix, $a$ uses a mixture per teacher input channel, while $\beta$,
$A_{\log}$, and the $\mathrm{dt}$ bias use per-target-head mixtures.
Convolutions use mixtures indexed by kernel tap and teacher output channel,
shared across its repeated target channels; the output norm uses mixtures
per teacher within-head channel, likewise shared across repeated slots.
All these mixture parameters are shared across participating layers.

\subsection{Mapping Optimization and Parameter Accounting}
\label{app:linmorph_optimization}

The teacher and native GDN template are frozen during mapping; differentiation
through the mapped block updates only the transformation.
The base step trains mixtures, channel adapters, bridges, and auxiliary
mappings without a residual. The second step freezes the complete base and
trains only native-shape residuals for $q,k,v,o,g,\beta$.
Both use Equation~\ref{eq:linmorph_map_loss}, with no relative-error
normalization or additional mapping regularizer.
The default mapping configuration includes all 30 layers in each update.
A long-video configuration instead uses cyclic, rank-strided layer
minibatches; this is a deterministic schedule, not independent random
layer sampling. Timesteps are sampled from the scheduler's discrete training
grid; uniform grid-index sampling need not be uniform in noise level.

An optional low-rank correction replaces $\Delta\mathbf{W}_j$ with
$\mathbf{U}_j\mathbf{V}_j^\top$, where
$\mathbf{U}_j\in\mathbb{R}^{d_{\mathrm{out},j}\times r}$ and
$\mathbf{V}_j\in\mathbb{R}^{d_{\mathrm{in},j}\times r}$.
One factor is Gaussian-initialized with standard deviation $0.02$ and the
other is zero; there is no additional $1/r$ scaling.
As with full corrections, these factors are shared across layers.

Parameter counts must include more than source-mixture coefficients.
In the specified FullMix configuration, the five dense channel adapters
contain $75{,}497{,}472$ parameters, the two bridges contain $9{,}437{,}184$,
and the main/auxiliary mixtures contain $52$, totaling $84{,}934{,}708$
trainable base parameters.
The subsequent full-residual step trains $151{,}068{,}672$ parameters while
freezing the base. The combined transformation contains $236{,}003{,}380$ learned parameters
and is shared across depth; it is not a per-layer mapping of that size.
Only one of the two parameter subsets is trainable in each fitting step.
For comparison, one native GDN module contains $151{,}241{,}264$ parameters;
its two bridges bring the complete replacement to $160{,}678{,}448$.
The base mapper is therefore $43.8\%$ smaller than a single native GDN,
or $47.1\%$ smaller than a complete replacement with bridges.
The residual-only fitting step is approximately the size of one native GDN;
it is not a second jointly trained copy of the base mapper.
Table~\ref{tab:linmorph_parameter_comparison} compares the parameters optimized
in each mapping or progressive-recovery phase.

\begin{table}[t]
    \centering
    \small
    \caption{Parameter-count comparison for the specified architecture.
    Mapping rows count the parameters trainable in each separate fitting step.
    Progressive recovery trains all selected GDN modules and their bridges.
    Full-DiT fine-tuning additionally updates the remaining model components.
    Counts are not measurements of wall-clock time.}
    \label{tab:linmorph_parameter_comparison}
    \begin{tabular}{lr}
        \hline
        Parameter set & Parameters (M)\\
        \hline
        Shared base mapper & 84.93\\
        Shared residual refinement (base frozen) & 151.07\\
        \hline
        Single native GDN (reference) & 151.24\\
        Single GDN with bridges (reference) & 160.68\\
        \hline
        7 GDN modules with bridges & 1,124.75\\
        14 GDN modules with bridges & 2,249.50\\
        21 GDN modules with bridges & 3,374.25\\
        \hline
    \end{tabular}
\end{table}

For clarity, the combined stored base and residual parameters total
$236.00$M, which exceeds a single replacement's parameter count.
The single-module comparison above refers to the base mapper and the
separate residual-fitting step, not to their combined stored size.
The shared mapping nevertheless remains substantially smaller than
multi-module or full-model recovery.
Materialization removes online source mixing and weight generation, but
does not remove the native GDN parameters or, in the default runtime, the
hidden-width bridges.

\subsection{Progressive Replacement and Recovery Details}
\label{app:linmorph_recovery}

With zero-based block indices, the nested replacement sets are
\begin{equation}
\begin{aligned}
    \mathcal{R}_0&=\varnothing,\\
    \mathcal{R}_1&=\{0,4,8,12,16,20,24\},\\
    \mathcal{R}_2&=\{0,1,4,5,8,9,12,13,16,17,20,21,24,25\},\\
    \mathcal{R}_3&=\{0,1,2,4,5,6,8,9,10,12,13,14,16,17,18,\\
                   &\hspace{1.5em}20,21,22,24,25,26\}.
\end{aligned}
\end{equation}
For stage $m$, a layer in $\mathcal{R}_{m-1}$ inherits its preceding checkpoint,
whereas a layer in $\mathcal{R}_m\setminus\mathcal{R}_{m-1}$ receives the
corresponding materialized initialization. All replacement attention modules at layers in $\mathcal{R}_m$,
including their bridges, are trainable during that stage; the remaining
student components are frozen. The teacher is always the original softmax
model.

Prediction distillation uses an elementwise L1 loss with unit coefficient,
without a task-loss term, trajectory loss, or scheduler loss weighting.
Teacher and student use the same conditioning and classifier-free
guidance (CFG) schedule.
For the seven-block stage and the one-shot recovery baseline,
CFG is set to five for the first 500 optimizer updates and to one
thereafter.
The subsequent fourteen- and twenty-one-block stages use CFG one
throughout.One teacher prediction is evaluated for each sampled noisy input;
no multi-step teacher trajectory is required.
After replacement reaches $\mathcal{R}_{m_\star}$, the final teacher-free
phase trains the entire hybrid DiT, including its retained softmax blocks,
while freezing the VAE and text encoder.
The 14-GDN setting uses $m_\star=2$ and 4,000 updates in this final phase;
the 21-GDN setting uses $m_\star=3$. The 4,000-update count is not the total
mapping-plus-recovery budget.

\paragraph{Compute and scope.}
For fixed head dimensions, each replaced recurrent attention operator has
linear sequence-length scaling and a fixed-size recurrent state.
The retained softmax layers mean that the full hybrid backbone is not
asymptotically linear in sequence length.
Training-memory use and wall-clock speed also depend on kernel realization,
projections, intermediate activations, and retained modules; they should be
measured separately.
Comparisons of initialization quality should report mapping cost, recovery
updates, and end-to-end conversion cost, and distinguish equal-recovery-budget
comparisons from equal-total-budget comparisons.

\section{Experimental Protocols and Additional Results}
\label{app:experimental_protocol}

\subsection{Recovery Protocols and Checkpoint Accounting}

\paragraph{Backbone-specific recovery objectives.}
The Wan2.1 initialization ablation uses teacher-prediction L1 distillation
without a task loss. Its matched historical CFG schedule uses scale five
for the first 500 updates and scale one thereafter. All three
initialization arms follow this schedule. The Wan2.2 ablation instead uses
\begin{equation}
    \mathcal{L}_{\mathrm{Wan2.2}}
    =w(t)\left[
      \operatorname{MSE}(v_S,v^\star)
      +0.05\operatorname{MSE}(v_S,\operatorname{sg}(v_T))
    \right],
\end{equation}
where $v^\star$ is the flow-matching target, $w(t)$ is the scheduler's
training weight, and $\operatorname{sg}$ denotes a detached teacher output.
Wan2.2 recovery uses OpenVid clips at $81\times736\times1280$,
with CFG one; functional evaluation uses HQ videos at
$81\times480\times832$.
Within each backbone, all arms update the same selected GDN modules and
their bridges while freezing the remaining DiT, VAE, and text encoder.

\begin{table}[t]
\centering
\small
\caption{Recovery settings for the initialization ablations. These are not
the mapper-fitting or final teacher-free fine-tuning settings.}
\label{tab:recovery_protocol}
\begin{tabular}{lll}
\toprule
Setting & Wan2.1 & Wan2.2 \\
\midrule
Source backbone & T2V-1.3B & TI2V-5B, T2V mode \\
Replacement blocks & $\{0,4,8,12,16,20,24\}$ & $\{0,5,10,15,20,25\}$ \\
Recovery data & HQ videos & OpenVid \\
Training clip shape & $81\times480\times832$ & $81\times736\times1280$ \\
Training devices & 8 GPUs & 8 GPUs \\
Per-device batch & 1 & 1 \\
Gradient accumulation & 2 & 4 \\
Effective batch & 16 & 32 \\
Learning rate & $5\times10^{-6}$ & $5\times10^{-6}$ \\
Warmup updates & 500 & 500 \\
Training seed & 20260731 & 20260731 \\
Evaluation seed & 20260921 & 20260922 \\
\bottomrule
\end{tabular}
\end{table}

\paragraph{Manual initialization.}
The manual baseline applies deterministic dimension alignment to the
pretrained query, key, value, and output projections, using the
backbone-specific subhead/repetition construction. It initializes the
$g$, $\beta$, and $a$ projection matrices to zero and the raw
short-convolution kernels with a unit last tap. Other decay and
normalization parameters retain their seeded base initialization.
Zero $\beta$ logits imply a sigmoid value of $0.5$, not a zero update gate.
The random baseline uses native seeded initialization of the target
replacement. The comparisons evaluate complete initialization recipes,
rather than isolating the act of copying $Q,K,V,O$ alone.

\paragraph{Mapper variants.}
Wan2.1 uses the dense-adapter FullMix morphism with a separately fitted
full-matrix residual correction described in Section~\ref{sec:cross_parameter}.
Wan2.2 uses a separately trained shared four-source mapper with
repetition-based dimension alignment and rank-eight additive corrections.
Both variants construct layer-specific target weights from a transformation
shared across source layers; each backbone uses a mapper fitted to its own
source weights and target dimensions.

\paragraph{Progressive and direct schedules.}
The progressive model uses 1,500 updates at each of seven, fourteen, and
twenty-one GDN blocks. Its last checkpoint is named after 3,000
\emph{last-stage microsteps}, but its total recovery budget is
$3\times3{,}000/2=4{,}500$ updates. The direct model uses
$9{,}000/2=4{,}500$ updates at twenty-one blocks.
The seven-block stage and the start of direct recovery both use CFG five
for 500 updates, followed by CFG one. The later fourteen- and twenty-one-
block progressive stages use CFG one throughout and restart the 500-update
learning-rate warmup. Optimizer/scheduler restarts and the number of
trainable modules are therefore part of the whole-recipe comparison.
Total optimizer updates are matched; per-layer exposure and total
training FLOPs differ because the number of replacements changes by stage.

\paragraph{Checkpoint accounting.}
Ablation budgets throughout the main paper count optimizer updates rather than
microsteps. The 14-GDN generation model is evaluated after 4,000 optimizer
updates of teacher-free full-model fine-tuning following mapping and
progressive distillation. The 21-GDN quality and efficiency evaluations use
the same final-recovery checkpoint at microstep 40,000, corresponding to
20,000 optimizer updates with gradient accumulation factor two. These counts
refer to the final recovery stage and exclude mapper fitting and preceding
progressive-distillation stages.

\subsubsection{Additional Recovery Curves}
\label{app:recovery_curves}

\begin{figure}[tbp]
    \centering
    \includegraphics[width=0.62\textwidth]{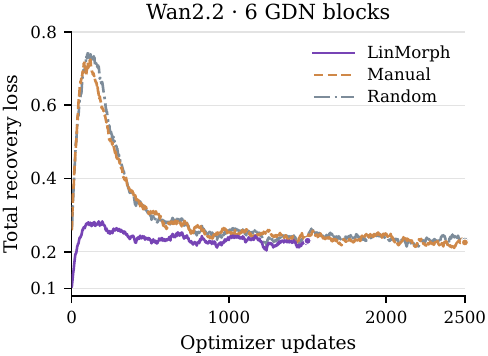}
    \caption{\textbf{Wan2.2 recovery with six GDN blocks.}
    Total training recovery loss with EMA-100 smoothing.
    All methods use the same backbone-specific recovery objective.}
    \label{fig:recovery_curves}
\end{figure}

Figure~\ref{fig:recovery_curves} compares the training dynamics of
the three initialization recipes on Wan2.2 with six GDN replacements.

\subsection{Evaluation Metrics and Statistical Scope}
\label{app:evaluation_protocol}

We measure functional fidelity using normalized mean squared error (NMSE) and
cosine similarity,
\begin{equation}
\begin{aligned}
\operatorname{NMSE}(\mathbf{y}_S,\mathbf{y}_T)
&=
\frac{\|\mathbf{y}_S-\mathbf{y}_T\|_F^2}
{\max(\|\mathbf{y}_T\|_F^2,10^{-12})},\\
\operatorname{Cos}(\mathbf{y}_S,\mathbf{y}_T)
&=
\frac{
\langle
\operatorname{vec}(\mathbf{y}_S),
\operatorname{vec}(\mathbf{y}_T)
\rangle
}{
\max(\|\mathbf{y}_S\|_F,10^{-12})
\max(\|\mathbf{y}_T\|_F,10^{-12})
}.
\end{aligned}
\end{equation}
Metrics are computed per example and then averaged.

Teacher-forced evaluation measures individual replacements using the
corresponding teacher hidden states as input. Depth-rollout evaluation instead
propagates student hidden states through a complete DiT forward pass and
measures both replaced-block outputs and the final velocity prediction.

Initialization and recovery-schedule ablations use a fixed 64-video HQ subset
at $81\times480\times832$ and diffusion timesteps
$\{100,300,500,700,900\}$. The 64-video evaluation subset uses indices
8192--8255. Within each ablation suite, record keys comprising video
index, timestep, block index, and evaluation scope match across candidates.
For $K$ replaced blocks, attention and depth-rollout block metrics each
average $64\times5\times K$ records; final output metrics average
$64\times5=320$ records. These are single-forward comparisons at sampled
noise levels, not diffusion-trajectory or generated-video measurements.

Mapper fitting uses a configured training prefix of 8,192 videos.
The fixed evaluation indices lie outside that prefix, but the loader's
decode-failure resampling may draw outside it, and exclusion from subsequent
Wan2.1 recovery data is not enforced.
Wan2.2 recovery uses OpenVid and evaluation uses HQ videos; cross-manifest source-video deduplication has not been performed. We accordingly describe these as fixed-subset functional comparisons.

Each configuration uses one training run.
Mapper-intervention intervals are obtained by averaging paired differences
over timesteps and layers within each of eight videos, then bootstrapping
video clusters 10,000 times.
The statistic is the ratio of mean intervention NMSE to mean unmodified
NMSE, minus one. Intervals describe video-sample variation for a fixed
trained mapper.

\subsection{Progressive Recovery: Complete Prediction Diagnostics}
\label{app:progressive_diagnostics}

\begin{table}[t]
\centering
\small
\setlength{\tabcolsep}{4pt}
\caption{\textbf{Complete diagnostics before final joint fine-tuning.}
Both 21-GDN models use 4,500 distillation updates. Block metrics measure
depth-rollout outputs at replaced layers; output metrics measure the
final DiT velocity prediction on the same fixed evaluation inputs.}
\label{tab:progressive_full}
\begin{tabular}{lrrrr}
\toprule
Schedule & \shortstack{Block\\NMSE $\downarrow$} & \shortstack{Block\\Cos.\ $\uparrow$} & \shortstack{Output\\NMSE $\downarrow$} & \shortstack{Output\\Cos.\ $\uparrow$} \\
\midrule
One-shot & 0.18406 & 0.90358 & 0.03415 & 0.98260 \\
Progressive & 0.17931 & 0.90728 & 0.03868 & 0.98026 \\
\bottomrule
\end{tabular}
\end{table}

The measurements in Table~\ref{tab:progressive_full} are taken immediately
after prediction distillation, before teacher-free joint fine-tuning.
Progressive conversion improves block alignment, while the one-shot model
has lower final-output NMSE at this stage.
Final joint fine-tuning optimizes the entire DiT under the flow-matching
objective so that intermediate representations and the denoising output
can adapt together.
The schedule comparison here covers the distillation stage; a matched
post-fine-tuning schedule comparison is not included.

\subsection{Additional Functional Diagnostics}
\label{app:additional_fidelity}

\paragraph{Independent training runs and hardware platforms.}
The pre-recovery initialization results in
Table~\ref{tab:init_fidelity} and the transformation-design ablation
in Section~\ref{sec:transformation_design} come from two independently
executed training and evaluation runs.
The former uses NVIDIA A800 GPUs, whereas the latter uses MetaX GPUs.
Both runs follow the same training and evaluation recipe for the
corresponding configurations, but their mapper parameters are fitted
separately.
Accordingly, matching configuration names across these experiments
refer to the same method specification, not to an identical trained
checkpoint.
Each table reports measurements from its own run rather than an
average across hardware platforms.

\paragraph{Pre-recovery fidelity protocol.}
Table~\ref{tab:init_fidelity} evaluates 64 QWorld videos from the
HQ evaluation data at $17\times256\times448$, using five diffusion
timesteps and all 30 source blocks.
Random initialization, manual copying, the base morphism, and the
residual-refined morphism receive identical cached video frames,
text conditions, and noise seeds.
All measurements are obtained before downstream recovery.

\begin{table}[t]
\centering
\small
\setlength{\tabcolsep}{4.5pt}
\caption{\textbf{Teacher-forced attention diagnostics for initialization.}
Rows shared with Table~\ref{tab:initialization} use the same checkpoints. An additional 1,500-update LinMorph checkpoint is included for Wan2.2.
A norm ratio close to zero indicates an attenuated student attention output;
low NMSE alone need not imply good directional agreement.}
\label{tab:attention_diagnostics}
\begin{tabular}{llrrrr}
\toprule
Backbone & Initialization & Updates & Attn.\ NMSE $\downarrow$ & Attn.\ Cos.\ $\uparrow$ & Norm ratio \\
\midrule
Wan2.1 & LinMorph & 1500 & 0.43849 & 0.75766 & 0.86764 \\
Wan2.1 & Manual & 1500 & 1.02142 & 0.26047 & 0.36950 \\
Wan2.1 & Random & 1500 & 1.10699 & 0.08895 & 0.26422 \\
Wan2.1 & Random & 3000 & 1.09900 & 0.15997 & 0.39579 \\
Wan2.2 & LinMorph & 1250 & 0.96955 & 0.48265 & 0.92147 \\
Wan2.2 & LinMorph & 1500 & 1.00650 & 0.47709 & 0.94721 \\
Wan2.2 & Manual & 1250 & 0.94925 & 0.25321 & 0.28571 \\
Wan2.2 & Manual & 2500 & 0.95883 & 0.31231 & 0.44852 \\
Wan2.2 & Random & 1250 & 1.03219 & 0.06895 & 0.22032 \\
Wan2.2 & Random & 2500 & 1.11141 & 0.13078 & 0.39946 \\
\bottomrule
\end{tabular}

\end{table}

Table~\ref{tab:attention_diagnostics} reports teacher-forced attention
metrics omitted from the main initialization table. They need not rank
models identically to depth-rollout or final-output metrics. For example,
Wan2.2 manual initialization at 1,250 updates obtains slightly lower
attention NMSE than LinMorph, but has much lower cosine similarity and
output norm ratio. An attenuated prediction can reduce squared error
without reproducing the teacher's direction. This motivates reporting
cosine similarity alongside NMSE and evaluating the composed model.

\paragraph{Degenerate fixed-copy sanity check.}
The zero-recovery mapping-analysis baseline initializes unmatched output
gate parameters to zero. Its attention output is zero (NMSE one and cosine zero), while the
surrounding block retains its residual stream.
Removing all unmatched routes from the fitted mapper also produces
zero attention output.
We therefore use route removal and source permutation with the other
learned components retained to test the fitted mapper's source assignments
in Section~\ref{sec:learned_structure}. Coefficient magnitudes visualize the
mixing rule, whereas these interventions test its functional effect.

\subsubsection{Cross-Scale Reuse of the Morphism}
\label{app:cross_scale}
Because the morphism operates on model weights rather than video tokens,
its parameterization does not depend on video length or spatial resolution
at a fixed architecture. This allows the same fitted morphism to be
reused at larger video scales.

The cross-scale results in Table~\ref{tab:shape_transfer}
use the original OpenVid evaluation subset, whereas
Table~\ref{tab:init_fidelity} reports the QWorld evaluation.
The small-scale measurements in these two tables therefore
refer to different evaluation inputs.

Both the base and residual-refined mappers are fitted at
$17\times256\times448$ and evaluated at this scale and at
$81\times480\times832$, without further optimization.
The evaluations use the same 64 video indices, five timesteps,
and 30 source blocks, yielding 9,600 teacher-forced local
comparisons per setting.
Each replacement receives the teacher's incoming hidden state,
so these measurements isolate local functional alignment.

\begin{table}[t]
\centering
\small
\setlength{\tabcolsep}{4.5pt}
\caption{\textbf{Reuse across video shapes without mapper refitting.}
Both variants are fitted at $17\times256\times448$; each is then evaluated
at both sizes. Each row averages 9,600 teacher-forced local comparisons
(64 videos, five timesteps, 30 blocks). Residual denotes the separately
fitted native-shape correction, not post-transfer optimization.}
\label{tab:shape_transfer}
\begin{tabular}{llrrrr}
\toprule
Evaluation size & Mapper & \shortstack{Attention\\NMSE $\downarrow$} & \shortstack{Attention\\Cos.\ $\uparrow$} & \shortstack{Block\\NMSE $\downarrow$} & \shortstack{Block\\Cos.\ $\uparrow$} \\
\midrule
$17\times256\times448$ & Base & 0.29951 & 0.83543 & 0.01242 & 0.99379 \\
$17\times256\times448$ & Base + residual & 0.27237 & 0.85161 & 0.01121 & 0.99439 \\
$81\times480\times832$ & Base & 0.34784 & 0.80592 & 0.01549 & 0.99231 \\
$81\times480\times832$ & Base + residual & 0.32078 & 0.82269 & 0.01418 & 0.99296 \\
\bottomrule
\end{tabular}

\end{table}

Table~\ref{tab:shape_transfer} shows that residual refinement
improves both attention- and block-level alignment at both scales.
At the fitting scale, it reduces block NMSE from $0.01242$
to $0.01121$.
At the larger scale, it reduces block NMSE from $0.01549$
to $0.01418$ and increases block cosine similarity from
$0.99231$ to $0.99296$.
The corresponding attention NMSE decreases from $0.34784$
to $0.32078$.
These improvements are obtained by reusing the same fitted
transformation and correction, rather than optimizing them
again at the larger scale.

\begin{figure}[t]
\centering
\includegraphics[width=\linewidth]{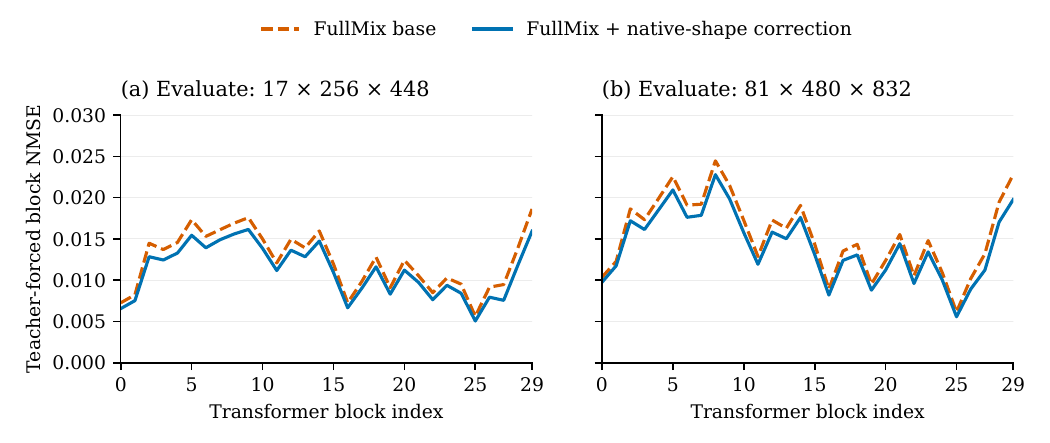}
\caption{\textbf{Layerwise local alignment across video sizes.}
The same low-resolution-trained base and residual-refined mappers are
evaluated at two video sizes. Block indices are zero-based; all 30 source
layers participated in mapper fitting. Lines connect measured layers and
are not uncertainty estimates.}
\label{fig:layerwise_shape_transfer}
\end{figure}

Figure~\ref{fig:layerwise_shape_transfer} complements the
aggregate results with block-NMSE profiles across all 30 layers.
Together, the table and layerwise profiles characterize how
the learned initialization transfers across video scales.

\paragraph{Mapping cost.}
The successful low-resolution FullMix base and residual runs each use
4,096 updates on four A800-80GB GPUs, batch one per device and no gradient
accumulation. Successful launcher-to-shutdown logs imply approximately
5.87 elapsed hours, or 23.48 GPU-hours, for both steps together; this
includes startup, I/O, and checkpointing, but excludes unsuccessful
earlier attempts. The stored base and residual together contain $236.00$M parameters;
the two fitting steps update $84.93$M and $151.07$M, respectively.
These shared fitting costs are reported separately from recovery-update
budgets. A larger-video fitting run uses layer minibatching and activation
checkpointing, so its elapsed time is not used as a matched-cost comparison.
The reported recovery-update reductions do not include mapper fitting.

\subsection{Cross-Layer Sharing and Generalization}
\label{app:cross_layer}

LinMorph uses a single transformation shared across participating layers,
rather than fitting an independent mapper for each replacement. This design
assumes that softmax-to-linear conversion contains reusable structure across
depth, even though each layer provides different pretrained source weights.
We examine this assumption by comparing mappings shared over one, six, and all
30 layers under the same data and fitting objective.

\begin{table}[tbp]
    \centering
    \caption{\textbf{Cross-layer sharing and generalization.}
    Block NMSE and cosine are averaged over five evenly spaced evaluation
    blocks (0, 6, 12, 18, and 24). For the one-layer setting, five
    separate mappers are each trained on
    one corresponding block; for the six-layer setting, five separate
    mappers are trained on consecutive groups $[0,5]$, $[6,11]$, $[12,17]$,
    $[18,23]$, and $[24,29]$. Each mapper is evaluated at all five evaluation
    blocks. The table then reports the equal-weight mean across the five
    mappers. The
    30-layer setting trains one mapper on all layers and evaluates it at the
    same five blocks.}
    \label{tab:cross_layer}
    \small
    \setlength{\tabcolsep}{6pt}
    \begin{tabular}{lcc}
        \toprule
        Layers sharing a mapper & Block NMSE $\downarrow$ & Block Cos. $\uparrow$ \\
        \midrule
        1 layer (independent mappers) & 0.20388 & 0.94327 \\
        6 layers (group-shared mappers) & 0.20734 & 0.94558 \\
        30 layers (one shared mapper) & \textbf{0.01560} & \textbf{0.99221} \\
        \bottomrule
    \end{tabular}
\end{table}

As shown in Table~\ref{tab:cross_layer}, the 30-layer shared mapper achieves
substantially better average alignment than the one-layer and six-layer
settings. Although both localized settings fit their corresponding blocks
slightly better, their alignment degrades on held-out anchors, so the local
fit gains do not persist in the cross-layer average. This pattern suggests
that learning across the full depth captures more reusable structure than
fitting mappers to individual layers or local groups, supporting a
full-depth shared mapper as our default strategy.

\subsection{Benchmark Protocols and Reference Sources}
\label{app:benchmark_details}

\paragraph{Sampling and checkpoint consistency.}
The LinMorph quality evaluations generate 81 frames at $480\times832$
and 16 fps, using the official Wan UniPC sampling configuration with
50 steps, CFG six, and flow shift eight.
Both benchmarks use microstep 8,000 for the 14-GDN model and microstep
40,000 for the 21-GDN model, corresponding to 4,000 and 20,000
final-recovery optimizer updates, respectively.

VBench contains 944 unique prompts with five samples per prompt,
for 4,720 staged videos, and uses the prescribed
temporal-flickering filter.
The 14-GDN run completes all 16 dimensions with no missing videos.
VBench-2.0 uses official Wan-augmented prompts checked against
original metadata, three samples per ordinary prompt and twenty
for diversity, for 3,860 videos per model.
Both configurations complete all 18 dimensions without recorded
evaluation errors.
Table~\ref{tab:vbench_totals} reports the respective full-suite
aggregates; the eight-dimension panel is not used to reconstruct
a VBench total.

\paragraph{Prompt correction for the starred 21-GDN total.}
For temporal flickering, 73 prompts retain the original
Wan-enhanced filtered videos.
The phone-booth and Petra prompts instead use the official
original positive prompts, with independently generated candidates
and the unchanged official RAFT acceptance procedure.
The resulting aggregate is recorded as a generation-variant result
and marked with $^*$ in Table~\ref{tab:vbench_totals}.
The eight dimensions in Table~\ref{tab:linvideo} exclude temporal
flickering and are unaffected by this correction.
The 14-GDN evaluation uses the standard prompt policy.

\paragraph{Aggregate reference sources.}
The source-backbone and efficient-inference reference scores come
from Table~3 of \citet{korzhenkov2026pyramidalwan}, including
the two- and one-step Wan-Adv/Wan-DMD configurations and
the selected PyramidalWan-DMD-PT$^\star$ model with a
2--2--1 sampling schedule.
These recent Wan-based methods provide context for alternative
routes to efficient generation.
Their sampling schedules differ from our 50-step
attention-converted models, as indicated in the table.

CogVideoX-2B, VideoREPA-2B, and MoAlign-2B use the jointly reported
VBench and VBench-2.0 evaluations in Table~3 of
\citet{bhowmik2026moalign}; VideoREPA is their reimplementation.
All published scores retain their reported precision.
The table presents selected reference systems rather than a
matched-training- or matched-inference-budget comparison.

\paragraph{LinVideo comparison.}
The dimension-wise reference is the LinVideo row in the
Wan2.1-1.3B section of Table~1 of \citet{huang2025linvideo},
corresponding to 16 converted attention blocks.
We include all eight reported dimensions:
imaging and aesthetic quality, motion smoothness and dynamic degree,
background and subject consistency, scene, and overall consistency.
Both systems use Wan2.1-1.3B and 480p video at 16 fps.
LinVideo reports unaugmented prompts and CFG five,
whereas LinMorph uses Wan-enhanced prompts and CFG six.
The comparison describes the evaluated generation systems;
the controlled ablation in Section~\ref{sec:init_ablation}
separately tests initialization.

\paragraph{Reference provenance.}
Table~\ref{tab:vbench_totals} consolidates full-suite VBench and
VBench-2.0 scores reported across several recent works. Because these systems
differ in model size, training data, sampling schedule, and inference budget,
the table is intended to provide broad generation-quality context rather than
a strictly matched comparison.

Superscript $^{a}$ denotes scores taken from Table~3 of
\citet{korzhenkov2026pyramidalwan}. This source provides the reported results
for Wan2.1-1.3B, Wan-Adv, Wan-DMD, and PyramidalWan-DMD-PT. We retain the
sampling configurations associated with the reported scores, including the
published PyramidalWan-DMD-PT$^{\star}$ variant with a $2$--$2$--$1$
sampling schedule.

Superscript $^{b}$ denotes scores reported in Table~3 of
\citet{bhowmik2026moalign}. These include CogVideoX-2B, MoAlign-2B, and the
VideoREPA-2B result reproduced in that work.

Superscript $^{c}$ denotes benchmark scores reported or collected by
\citet{ghafoorian2025attention} under the evaluation protocol used in that
work. In Table~\ref{tab:vbench_totals}, this source is used for
CogVideoX-1.5-5B, Open-Sora V1.2, LTX-Video, SnapGen-V, and M4V.
For original method descriptions, see
VideoCrafter2~\citep{chen2024videocrafter2},
Open-Sora~\citep{zheng2024opensora},
LTX-Video~\citep{hacohen2024ltxvideo},
SnapGen-V~\citep{wu2025snapgenv}, and
M4V~\citep{huang2026m4v}.
The superscripts in Table~\ref{tab:vbench_totals} identify the sources
of the benchmark scores, rather than the original method papers.

Superscript $^{d}$ denotes the VideoCrafter2 VBench score from the official
VBench leaderboard~\citep{huang2024vbench}; no comparable full-suite
VBench-2.0 score is available in the sources used here.

Superscript $^{e}$ identifies results sourced from the corresponding
method papers.
The HLA-3F-R1-10 result is from
\citet{ackermann2026hla}.
For Attention Surgery, we use the published CVPR version of
\citet{ghafoorian2025attention}.
The $15\times R2$ VBench and VBench-2.0 scores are taken
from its Tables~1 and~4, respectively.
The $20\times R4$ VBench score marked with $^{\dagger}$
is approximately read from the $480\times832$ panel of
its Figure~5 and reported to one decimal place.
HLA uses Hadamard linear attention, M4V uses a Mamba/SSM backbone,
and Attention Surgery uses hybrid linear--softmax attention.
These methods provide related linear-attention, state-space, and
hybrid video-generation references.

Finally, all LinMorph scores are obtained from our own evaluation.
The $^{*}$ marker on LinMorph-21 denotes the temporal-flickering prompt
correction described above; it is unrelated to the $\star$ used in the
published PyramidalWan-DMD-PT model name.

\subsection{Runtime Measurement Details}
\label{app:runtime_details}

\begin{table}[t]
\centering
\small
\setlength{\tabcolsep}{4.5pt}
\caption{\textbf{Full-generation latency and peak allocated memory.}
All runs use $480\times832$, batch one, BF16, 50 denoising steps, and CFG six.
Each latency is one timed generation after one excluded first generation.
Memory is PyTorch peak allocated memory in GiB, not total board memory.}
\label{tab:runtime_details}
\begin{tabular}{rrrrrr}
\toprule
Frames & Wan (s) & 21 GDN (s) & Speedup & Wan (GiB) & 21 GDN (GiB) \\
\midrule
81 & 166.9 & 213.2 & 0.78$\times$ & 15.50 & 28.40 \\
241 & 959.0 & 753.7 & 1.27$\times$ & 18.42 & 44.61 \\
321 & 1604.7 & 1094.0 & 1.47$\times$ & 19.87 & 52.72 \\
\bottomrule
\end{tabular}

\end{table}

\paragraph{Implementation and scopes.}
Each measurement uses one NVIDIA A800-SXM4-80GB, BF16, and batch size one.
The environment uses PyTorch 2.6.0 with CUDA 12.6, Triton 3.2.0,
FlashAttention 2.6.3, and FLA 0.3.2; TF32 is disabled.
The attention-module benchmark measures the trained layer-zero module,
including projections, gates, convolutions, recurrence, normalization,
and input/output bridges, on identical seeded synthetic hidden states.
It is not a recurrence-kernel-only measurement.
The backbone benchmark measures one denoiser forward on synthetic latents
and conditioning, excluding text encoding and VAE decoding; each timed
call contains one conditioning branch, not a full CFG sampling step.
Full-generation timing includes text encoding, all denoising steps,
tiled VAE decoding, and conversion to PIL frames, but excludes model
loading and MP4 encoding. It uses one fixed prompt and seed 42, with
50 steps, CFG six, and separate conditional/unconditional forwards.

\paragraph{Repetitions and aggregation.}
Module and backbone runs exclude an initial first call, perform five
additional warmup calls, then time twenty calls per model in each of
three paired rounds. Baseline and student are paired on the same physical
GPU. Each round forms the ratio of median baseline to median student
latency; the plotted point is the median of these three ratios and the
band is their interquartile range.
Full-generation runs exclude an initial complete generation and time
one further generation per shape and model. These full-generation points have one timed sample each.
Peak allocated memory includes resident model and input allocations;
it is not incremental attention workspace or total GPU-board memory.

\paragraph{Resource accounting and runtime configuration.}
Empty-card utilization/memory checks and memory-ledger guards passed.
Host/container PID-namespace mapping did not establish exclusive ownership.
All reported results use the original FLA implementation.
The 21-GDN denoiser contains 4,594,869,808 parameters; the pretrained
baseline contains 1,418,996,800.
At 321 frames, peak allocated full-generation memory is 52.72 and
19.87 GiB, respectively.
The latency comparison measures attention replacement, while mapped
initialization controls the training starting point.
At short sequences, projection and recurrent overhead can dominate;
the table and scaling figure include these points as well as longer videos.
The nine retained softmax blocks preserve a quadratic component in the
hybrid's total computation.
Generation-quality measurements use 81 frames; longer inputs are used
for the runtime scaling study.

\subsection{Qualitative Comparisons}
\label{app:qualitative_comparison}

\par\noindent
\begin{minipage}{\linewidth}
    \centering
    \includegraphics[
        width=\linewidth,
        keepaspectratio
    ]{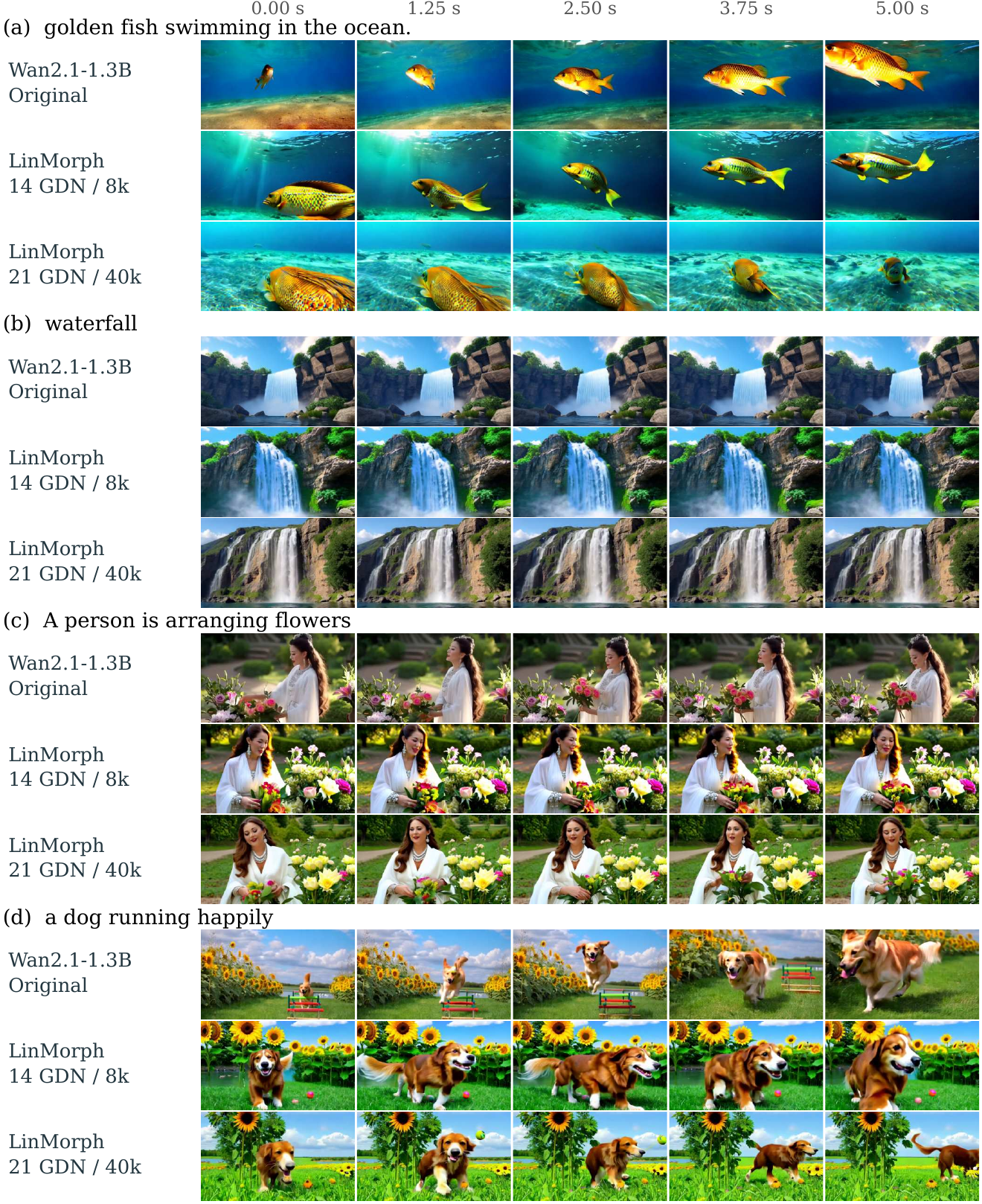}

    \captionsetup{hypcap=false}
    \captionof{figure}{
    \textbf{Qualitative comparison on four selected prompts.}
    Rows within each group show Wan2.1-1.3B, LinMorph-14,
    and LinMorph-21 using identical enhanced prompts,
    seeds, and sampling settings.
    Columns show matched timestamps; 8k and 40k denote
    fine-tuning microsteps.
    }
    \label{fig:qualitative_comparison}
\end{minipage}
\par

\end{document}